\documentclass{article}

    \usepackage[final]{neurips_2022}

\usepackage[utf8]{inputenc} 
\usepackage[T1]{fontenc}    
\usepackage{url}            
\usepackage{booktabs}       
\usepackage{amsfonts}       
\usepackage{nicefrac}       
\usepackage{microtype}      
\usepackage{xcolor}         
\usepackage[numbers]{natbib}

\usepackage{fleqn, tabularx}
\usepackage{multirow}
\usepackage{mathtools}
\usepackage[export]{adjustbox}
\usepackage{amsmath}
\usepackage{amssymb}
\usepackage{colortbl}
\usepackage{wrapfig}
\usepackage{algorithm}
\usepackage[noend]{algorithmic}
\usepackage{xcolor}

\usepackage{mathrsfs}
\usepackage{chngcntr}
\usepackage[colorlinks=true,linkcolor=blue,citecolor=blue]{hyperref}
\usepackage[skins,theorems]{tcolorbox}
\usepackage{relsize}
\usepackage{todonotes}
\usepackage{derivative}

\usepackage{thmtools}
\usepackage{thm-restate}
\usepackage{pythonhighlight}

\usepackage{wrapfig}
\usepackage{pdfpages}

\lstnewenvironment{mypython}[1][]{\lstset{style=mypython,#1}}{}

\newcommand{\edit}[1]{ #1 }

\newcommand{\policy}{\pi}

\newcommand{\name}{\text{DASCO}} 

\usepackage{amsmath,amsfonts,bm}

\def\eqref#1{equation~\ref{#1}}

\def\1{\bm{1}}

\DeclareMathAlphabet{\mathsfit}{\encodingdefault}{\sfdefault}{m}{sl}
\SetMathAlphabet{\mathsfit}{bold}{\encodingdefault}{\sfdefault}{bx}{n}

\newcommand{\E}{\mathbb{E}}

\DeclareMathOperator*{\argmax}{arg\,max}
\DeclareMathOperator*{\argmin}{arg\,min}

\title{ Dual Generator Offline Reinforcement Learning }

\author{Quan Vuong$^{1}$, Aviral Kumar$^{1, 2}$, Sergey Levine$^{1,2}$,  Yevgen Chebotar$^{1}$ \vspace{0.3cm}\\
$^1$Google Research, $^2$UC Berkeley \\ 
}

\begin{document}

\maketitle

\begin{abstract}
In offline RL, constraining the learned policy to remain close to the data is essential to prevent the policy from outputting out-of-distribution (OOD) actions with erroneously overestimated values.
In principle, generative adversarial networks (GAN) can provide an elegant solution to do so, with the discriminator directly providing a probability that quantifies distributional shift. However, in practice, GAN-based offline RL methods have not performed as well as alternative approaches, perhaps because the generator is trained to both fool the discriminator and maximize return -- two objectives that can be at odds with each other. In this paper, we show that the issue of conflicting objectives can be resolved by training two generators: one that maximizes return, with the other capturing the ``remainder'' of the data distribution in the offline dataset, such that the mixture of the two is close to the behavior policy. 
We show that not only does having two generators enable an effective GAN-based offline RL method, but also approximates a support constraint, where the policy does not need to match the entire data distribution, but only the slice of the data that leads to high long term performance. We name our method \name, for \textbf{D}ual-Generator \textbf{A}dversarial \textbf{S}upport \textbf{C}onstrained \textbf{O}ffline RL. On benchmark tasks that require learning from sub-optimal data, \name{} significantly outperforms prior methods that enforce distribution constraint.

\end{abstract}

\vspace{-0.25cm}
\section{Introduction}
\vspace{-0.25cm}

Offline reinforcement learning (RL) algorithms aim to extract policies from datasets of previously logged experience. The promise of offline RL is to extract \textit{decision making engines} from existing data \cite{levine2020offline}. Such promise is especially appealing in domains where data collection is expensive or dangerous, but large amounts of data may already exists (e.g., robotics, autonomous driving, task-oriented dialog systems). Real-world datasets often consist of both expert and sub-optimal behaviors for the task of interest and also include potentially unrelated behavior corresponding to other tasks. While not all behaviors in the dataset are relevant for solving the task of interest, even sub-optimal trajectories can provide an RL algorithm with some useful information. In principle, if offline RL algorithms can combine segments of useful behavior spread across multiple sub-optimal trajectories together, the combined segments can then perform better than any behavior observed in the dataset. 

Effective offline RL requires estimating the value of actions other than those that were taken in the dataset, so as to pick actions that are better than the actions selected by the behavior policy. However, this requirement introduces a fundamental tension: the offline RL method must generalize to new actions, but it should not attempt to use actions in the Bellman backup for which the value simply cannot be estimated using the provided data. These are often referred to in the literature as out-of-distribution (OOD) actions~\citep{bear}. While a wide variety of methods have been proposed to constrain offline RL to avoid OOD actions~\citep{kostrikov2021offline, bcq, agarwal2019optimistic}, the formulation and enforcement of such constraints can be challenging, and might introduce considerable complexity, such as the need to explicitly estimate the behavior policy~\citep{brac} or evaluate high-dimensional integrals~\citep{kumar2020conservative}. Generative adversarial networks (GANs) in principle offer an appealing and simple solution: use the discriminator to estimate whether an action is in-distribution, and train the policy as the ``generator'' in the GAN formulation to fool this discriminator. Although some prior works have proposed variants on this approach~\citep{brac}, it has been proven difficult in practice as GANs can already suffer from instability when the discriminator is too powerful. Forcing the generator (i.e., the policy) to simultaneously \emph{both} maximize reward and fool the discriminator only exacerbates the issue of an overpowered discriminator.

We propose a novel solution that enables the effective use of GANs in offline RL, in the process not only mitigating the above challenge but also providing a more appealing form of support constraint that leads to improved performance. Our key observation is that the generative distribution in GANs can be split into \emph{two} separate distributions, one that represents the ``good parts'' of the data distribution and becomes the final learned policy, and an auxiliary generator that becomes the policy's complement, such that their mixture is equal to the data distribution. This formulation removes the tension between maximizing rewards and matching the data distribution perfectly: as long as the learned policy is within the \emph{support} of the data distribution, the complement will pick up the slack and model the ``remainder" of the distribution, allowing the two generators together to perfectly fool the discriminator. If however the policy ventures outside of the data support, the second generator cannot compensate for this mistake, and the discriminator will push the policy back inside the support. 
We name our method \name, for \textbf{D}ual-Generator \textbf{A}dversarial \textbf{S}upport \textbf{C}onstrained \textbf{O}ffline RL.

Experimentally, we demonstrate the benefits of our approach, \name{},
on standard benchmark tasks. For offline datasets that consist of a combination of expert, sub-optimal and noisy data,
our method outperforms distribution-constrained offline RL methods by a large margin.  


\vspace{-3pt}
\section{Related Work}
\vspace{-3pt}
Combining sub-optimal trajectories to obtain high-performing policies is a central promise of offline RL. During offline training, querying the value function on unseen actions often leads to value over-estimation and collapse in learning progress. To avoid querying the value functions on out-of-distribution actions, existing methods encourage the learned policies to match the distribution of the behavior policies. This principle has been realized with a variety of practical algorithms ~\citep{jaques2019way,brac,peng2019awr,siegel2020keep,brac,kumar2019stabilizing, kostrikov2021offline,kostrikov2021offlineb,wang2020critic,fujimoto2021minimalist, BAIL, furuta2022generalized, jang2022gptcritic, meng2022offline, daoudi2022density, liu2022robust}.
For example, by optimizing the policies with respect to a conservative lower bound of the value function estimate \cite{kumar2020conservative}, only optimizing the policies on actions contained in the dataset \cite{kostrikov2021offline}, or jointly optimizing the policy on the long-term return and a behavior cloning objective \cite{fujimoto2021minimalist}. While \textit{explicitly} enforcing distribution constraint by adding the behavior cloning objective allows for good performance on near-optimal data, 
this approach fails to produce good trajectories on sub-optimal datasets~\cite{kostrikov2021offline}.
Methods that \textit{implicitly} enforce distribution constraints, such as CQL and IQL, have seen more successes on such datasets.
However, they still struggle to produce near-optimal trajectories when the actions of the dataset generation policies are corrupted with noise or systematic biases (a result we demonstrate in Section~\ref{sec:exp}).

However, enforcing distribution constraints  to avoid value over-estimation may not be necessary. It is sufficient to ensure the learned policies do not produce actions that are too unlikely under the dataset generation policy. That is, it is not necessary for the learned policy to fully \emph{cover} the data distribution, only to remain in-support~\citep{kumar2019stabilizing,kumar_blog,levine2020offline,brac,zhou2020plas,chen2022latent}. Unfortunately, previous methods that attempt to instantiate this principle into algorithms have not seen as much empirical success as algorithms that penalize the policies for not matching the action distribution of the behavior policies. In this paper, we propose a new GAN-based offline RL algorithm whose use of dual generators naturally induce support constraint and has competitive performance with recent offline RL methods. In a number of prior works, GANs have been used in the context of imitation learning to learn from expert data~\citep{gail,infogail,intentiongan,zhihanliu}. In this work, we show that dual-generator GANs can be used to learn from sub-optimal data in the context of offline RL.


\edit{To enforce the support constraint, we use the discriminator score to measure how likely an action is under the behavior policy. To measure uncertainty, one could also use an ensemble of value functions, as is done in~\citep{EDAC, PBRL}. We include~\citep{EDAC} as a baseline in \autoref{tab:hetero_antmaze} and demonstrate the our method has significantly higher performance. Nevertheless, it would be interesting as future work to combine the discriminator score and an ensemble of value functions to devise a higher performing algorithm.}

\vspace{-1pt}
\section{Background}
\vspace{-1pt}
Let $\mathcal{M} = (\mathcal{S}, \mathcal{A}, P, R, \gamma)$ define a Markov decision process (MDP), where $\mathcal{S}$ and $\mathcal{A}$ are state and action spaces, $P: \mathcal{S} \times \mathcal{A} \times \mathcal{S} \rightarrow \mathbb{R}_{+}$ is a state-transition probability function, $R: \mathcal{S} \times \mathcal{A} \rightarrow \mathbb{R}$ is a reward function and $\gamma$ is a discount factor. Reinforcement learning methods aim at finding a policy $\pi(a | s)$ that maximizes the expected discounted reward $R(\tau) = \sum_{t=0}^T \gamma^t R(s_t,a_t)$  over trajectories $\tau = (s_0, a_0, \dots, s_T, a_T)$ with time horizon $T$ induced by the policy $\pi$. 

In this work, we concentrate on the offline or off-policy RL setting, i.e. finding an optimal policy given a dataset $\mathcal{D}$ of previously collected experience $\tau \sim \mathcal{D}$ by a behavior policy $\pi_\beta$. A particularly popular family of methods for offline learning are based on training a Q-function through dynamic programming using temporal-difference (TD) learning~\cite{watkins1992q,sutton1998}. Such methods train a Q-function to satisfy the Bellman equation:
 \[
 Q (s_t, a_t) = R(s_t, a_t) + \gamma \mathbb{E}_{a\sim\pi} [Q(s_{t+1}, a)].
 \]
In Q-learning, the policy is replaced with a maximization, such that $\pi(a | s) = \argmax_a Q_\theta(s,a)$, while actor-critic methods optimize a separate parametric policy $\pi_\phi(a | s)$ that maximizes the Q-function. In this work, we extend the Soft Actor-Critic (SAC) method~\cite{sac} for learning from diverse offline datasets.

Generative Adversarial Networks (GANs)~\cite{GoodfellowPMXWOCB14} enable modeling a data distribution $p_{\mathcal{D}}$ through an adversarial game between a generator $G$ and a discriminator $D$:
\begin{align}
    \min_{G} \max_{D}  \mathbb{E}_{ x \sim p_{\mathcal{D}}} [\log(D(x))] + \mathbb{E}_{z \sim p(z)} [\log(1-D(G(z)))]
    \label{eq:vanilla_gan}
\end{align}

For this two player zero-sum game, \cite{GoodfellowPMXWOCB14} shows that for a fixed generator $G$, the optimal discriminator is $ D^*_G(x) = \dfrac{ p_{\mathcal{D} } (x) }{ p_{\mathcal{D} } (x) + p_{G }(x) }$ and the optimal generator matches the data distribution $ p^*_g(x) = p_{\mathcal{D}} $.

GAN has been extended to the offline RL setting by interpreting the discriminator function as a measure of how likely an action is under the behavior policy, and jointly optimizing the policy to maximize an estimate of the long-term return and the discriminator function~\cite{brac}: 
\begin{align}
    \min_{\pi} \max_{D}  \mathbb{E}_{ s, a \sim p_{\mathcal{D}}} [\log(D(s,a))] + \mathbb{E}_{ s \sim p_{\mathcal{D}}, a \sim \pi(a|s) } [\log(1 - D(s,a))] - \mathbb{E}_{s \sim p_{\mathcal{D}}, a \sim \pi(a|s)} [ Q(s, a) ], \label{eq:brac}
\end{align}
where $Q(s,a)$ is trained via the Bellman operator to approximate the value function of the policy $\pi(a|s)$.
This leads to iterative policy evaluation and policy improvement rules for the actor and the policy~\cite{brac}. During the $k^{th}$ update step, given the most recent values for the policy $\pi^{k}$, the value function $Q^{k}$, and the discriminator $D^k$, we perform the following updates to obtain the next values for the value function and the policy:
\begin{align}
\begin{split}
\label{eq:brac_update}
    Q^{k+1} \leftarrow& \argmin_{Q} \E_{s, a, s' \sim \mathcal{D}}\left[ \left((R(s, a) +  \gamma \E_{a' \sim {\policy}^k(a'|s')}[Q^{k}(s', a')]) - Q_{target}(s, a)\right)^2 \right]\\ 
    \policy^{k+1} \leftarrow& \argmax_{\policy} \E_{s \sim \mathcal{D}, a \sim \policy^k(a|s)}\left[Q^{k+1}(s, a) + \log D^k(s,a) \right]
\end{split}
\end{align}
where the $\log D(a|s)$ term in the policy objective aims at regularizing the learnt policy to prevent it from outputting OOD actions. In practice, training the policy to maximize both the value function and discriminator might lead to conflicting objectives for the policy and thus poor performance on either objective.
This can happen when the data contains a mixture of good and bad actions. Maximizing the value function would mean avoiding low-reward behaviors. On the other hand, maximizing the discriminator would require outputting all in-distribution actions, including sub-optimal ones. Our approach alleviates this conflict and enables \textit{in support} maximization of the value function when learning from mixed-quality datasets.



\section{Dual-Generator Adversarial Support Constraint Offline RL}

We now present our algorithm, which uses a novel dual-generator GAN in combination with a weighting method to enable GAN-based offline RL that constrains the learned policy to remain within the support of the data distribution. We call our method \textit{Dual-generator Adversarial Support Constraint Offline RL (\name{})}. We will first introduce the dual-generator training method generically, for arbitrary generators that must optimize a user-specified function $f(x)$ within the support of the data distribution in Section~\ref{sec:dual_gen}. We will then show this method can be incorporated into a complete offline RL algorithm in Section~\ref{sec:update_rule} in combination with our proposed weighting scheme, and then summarize the full resulting actor-critic method in Section~\ref{sec:algo_summary}.

\subsection{Dual generator in-support optimization}
\label{sec:dual_gen}


\begin{wrapfigure}{R}{0.5\textwidth}
 \vspace{-4em}
  \begin{center}
    \includegraphics[width=\linewidth]{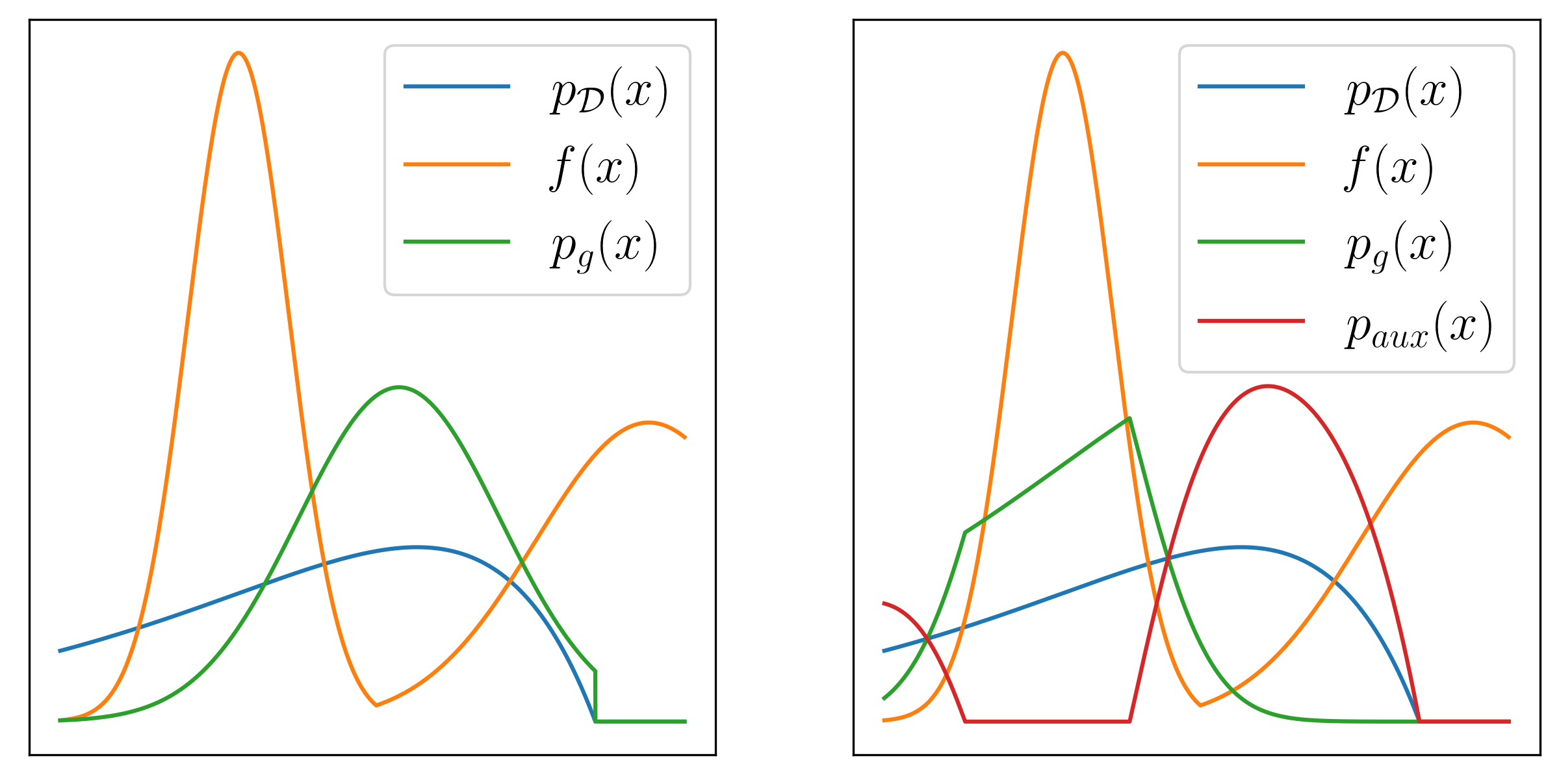}
    \vspace{-2em}
  \end{center}
  \caption{Visualizations to illustrate the benefit of \textit{dual} generators over single generator when maximizing a secondary objective $f(x)$ in the GAN framework. In both figures, $p_\mathcal{D}(x)$ is the data distribution. The x-axis is a one-dimensional sample space. {\bf Left:} In this figure, since there is only a single generator, the generator $G$ is trained to jointly maximize the objective $f(x)$ and matches the data distribution $p_\mathcal{D}(x)$. The distribution $p_G$ induced by the generator is thus not very good at either maximizing the objective $f(x)$ or matching the data distribution. {\bf Right:} In this figure, we have two generators, inducing two distributions $p_G$ and $p_{aux}$. By introducing the auxiliary generator $G_{aux}$ into the GAN framework, the primary generator can better maximize the objective $f(x)$ while staying within the support of the data distribution $p_\mathcal{D}$. The mixed distribution also perfectly matches the data distribution, i.e. $ \dfrac{p_g(x) + p_{aux}(x)}{2} = p_{\mathcal{D}} (x) $. Note that in these two figures, the primary generator aims to maximize $f(x)$ (instead of minimize) to allow for more intuitive interpretation. \edit{In Appendix~\ref{sec:1d_example}, we provide a numerical example to further illustrate the benefit of the dual generator technique on a 1D example.}}
  \label{fig:dual_gen_viz}
  \vspace{-5.5em}
\end{wrapfigure}

In this section, we will develop an approach for performing a joint optimization of adversarial and secondary objectives of the generator in a GAN framework, which we will then apply to offline RL. This is a necessary component for performing the joint optimization in Eq.~\ref{eq:brac} without introducing a conflict of these objectives. All proofs for theorems presented in this section are in Appendix A.

Let's consider a general objective that requires training a generator $G$ to fool the discriminator $D$ while also optimizing the expected value of some other function $f$:
\begin{align}
   \min_{G} \max_{D} \quad & \mathbb{E}_{ x \sim p_{\mathcal{D}}} [\log(D(x))] \nonumber \\
+ \hspace{0.3em} & \mathbb{E}_{z \sim p(z)} [\log(1-D(G(z)))] \nonumber \\
+ \hspace{0.3em} & \mathbb{E}_{z \sim p(z)} [ f ( G(z) ) ] \label{eq:joint_nodual}
\end{align}
where the first two terms are the same as the objective of the GAN formulation. We have also added an additional term $\mathbb{E}_{z \sim p(z)} [ f ( G(z) ) ]$, where $f$ is a mapping from the generator output to a scalar value. The third term represents a secondary objective that the generator should optimize.
\begin{restatable}{thm}{thmdisjoint}
\label{thm_disjoint}
The optimal generator of Eq.~\ref{eq:joint_nodual} induces a distribution $ p^*_g(x) = p_{\mathcal{D}} (x) \dfrac{ e^{- f(x) - \nu} }{ 2 - e^{ - f(x) - \nu } }$, where $\nu > 0$ is the Lagrange multiplier that ensures that $p^*_g(x)$ is normalized to 1.
\end{restatable}

We can see that by adding a secondary objective function for the generator, in general, the optimal generator does not attempt to match the data distribution $ p_{\mathcal{D}} (x) $ anymore, but instead tries to match the data distribution weighted by $ \dfrac{ e^{- f(x) - \nu } }{ 2 - e^{ - f(x) - \nu } } $. We expect that in such case, the discriminator clearly has an advantage in the two player zero-sum game and will be able to distinguish between real samples and sample generated by the generator.

To allow the generator to specialize in optimizing the secondary objective function, we propose to introduce a second auxiliary generator that matches the portion of the data distribution that is not well captured by the primary generator. Let $p_{mix} = \dfrac{ p_g + p_{aux} }{ 2 }$, consider the min-max problem:
\begin{align}
    \min_{G, G_{aux} } \max_{D}  \mathbb{E}_{ x \sim p_{\mathcal{D}} } [\log(D(x))] + \mathbb{E}_{ x \sim p_{mix} } [ \log(1-D(x))]  + \mathbb{E}_{ x \sim p_g } [ f ( x ) ], \label{eq:joint_gan}
\end{align}
where we mix samples from the primary generator $G$ and the auxiliary generator $G_{aux}$ to generate samples that can fool the discriminator. The mixing is indicated by the distribution $p_{mix}$ in the second term of Eq.~\ref{eq:joint_gan}. The first and third term of Eq.~\ref{eq:joint_gan} are the same as the objective in Eq.~\ref{eq:joint_nodual}.

We next theoretically demonstrate the benefit of adding the auxiliary generator to the GAN formulation with the following Theorem.

\begin{restatable}[Informal]{thm}{thmjointinsupportmax}
\label{thm_joint_in_support_max}
The primary generator $p_G$ performs in-support optimization of $f(x)$.
\end{restatable}

We first note that the optimal solution of the mixed distribution from Eq.~\ref{eq:joint_gan} is the real data distribution:
\begin{align}
\dfrac{ p^*_{aux} (x) + p^*_{g} (x) }{2} = p_{\mathcal{D} } (x)
\label{eq:joint_match_real}
\end{align}
Accordingly, the optimal auxiliary generator distribution can be expressed as 
\begin{align}
    p^*_{aux} (x) = 2 p_{\mathcal{D}} (x) - p^*_g(x) \label{eq:paux_dist}
\end{align}
Let $x_0$ to be the element inside the support of the data distribution $p_{\mathcal{D}}$ that minimizes $f$. That is: $$ x_0 = \underset{x \in \text{Supp}( p_{\mathcal{D}} ) }{\arg\min} f(x) $$
When optimizing the secondary objective $f(x)$, the primary generator will maximize the probability mass of in-support samples that maximize $f(x)$. However, Eq.~\ref{eq:paux_dist} introduces a constraint that enforces $ 2 p_{\mathcal{D}} (x) - p^*_g(x) \geq 0$ for $p^*_{aux} (x) \geq 0$ to remain a valid distribution. This leads us to conclude that the optimal primary generator $p^*_g$ assigns the following probability to $x_0$:
\begin{align}
    p^*_{g} (x_0) & = \begin{cases}
    2p_{\mathcal{D}} (x_0) \quad  \text{if} \quad 2p_{\mathcal{D}} (x_0) < 1 \\
    1 \quad \quad  \quad \quad \text{otherwise}
    \end{cases}
\end{align}
Interestingly, if the global optimum $x_0$ is not taking the full probability mass, the rest of the probability mass is redistributed to the next best in-support optimum, which we can define recursively:
\begin{align}
         \text{For} \,\, x_i \in \underset{x \in \text{Supp}( p_{\mathcal{D}} )  \setminus \{ x_j \}_{j=0}^{i-1} }{\arg\min} f(x), \,\,
    p^*_{g} (x_i) & = \begin{cases}
      2p_{\mathcal{D}} (x_i)\quad \quad \quad \quad \quad \text{if} \quad \sum_{j=0}^i p^*_g(x_j) < 1 \\
      1 -  \sum_{j=0}^{i-1} p^*_g(x_j) \quad \,\, \text{if} \quad \sum_{j=0}^i p^*_g(x_j) > 1 \\ 
      0 \quad \quad \quad \quad \quad \quad \quad \quad  \text{if} \quad \sum_{j=0}^{i-1} p^*_g(x_j) = 1 
    \end{cases}
    \label{eq:sol_3_cases}
\end{align}

\edit{ We provide more explanation for the solution in Eq.~\ref{eq:sol_3_cases}. 
In the first case, $p^*_{g} (x_i) = 2p_{\mathcal{D}} (x_i)$ if $\sum_{j=0}^i p^*_g(x_j) < 1$. 
That is, if the optimal solution for the primary generator $p^*_{g}$ \textit{can} assign the probability $2p_{\mathcal{D}} (x_i)$ to the $i^{th}$ in support minima of $f(x)$ without the total sum of probability assigned $\sum_{j=0}^i p^*_g(x_j)$ going over $1$, then the primary generator $p^*_{g}$ will assign the probability $2p_{\mathcal{D}} (x_i)$ to $x_i$. }

\edit{ In the second case, $p^*_{g} (x_i) = 1 -  \sum_{j=0}^{i-1} p^*_g(x_j)$ if $\sum_{j=0}^i p^*_g(x_j) > 1$. That is, if by assigning the probability $2p_{\mathcal{D}} (x_i)$ to the $i^{th}$ in support minima of $f(x)$, the total sum of probability assigned $\sum_{j=0}^i p^*_g(x_j)$ \textit{goes over} $1$, then the primary generator $p^*_{g}$ will assign the remaining probability $1 -  \sum_{j=0}^{i-1} p^*_g(x_j)$ to $x_i$. In the third case, the generator assigns probability $0$ to $x_i$ because all the probability has already been assigned. }

To summarize the benefit of dual generator, we note that by introducing an auxiliary generator and mixing it with the primary generator, not only does the optimal solution for the mixed distribution match the real data distribution, but also the primary generator can better optimize the secondary objective $f$ on the part of the domain of $f$ that is within the support of the data distribution $p_\mathcal{D}$. To better illustrate the benefit, we provide a visual explanation of the benefit in Figure~\ref{fig:dual_gen_viz}.

\subsection{Update rules for offline reinforcement learning}
\label{sec:update_rule}

We will now incorporate the dual-generator method to train policies for offline RL, based on optimizing the joint objective from Eq.~\ref{eq:joint_gan}. The updates for the actor and the critic are generally similar to Eq.~\ref{eq:brac_update}. However, simply combining Eq.~\ref{eq:joint_gan} and Eq.~\ref{eq:brac_update} can still allow the policy to exploit errors in the value function during the policy improvement step. We therefore augment the policy improvement step with an adaptive weight on the Q-value.

More concretely, as the policy improvement step samples actions from the current policy iterate $\policy^k$ to optimize the policy objective, ', as we observe in our experiments. To alleviate this issue, we use the probability assigned to the sampled actions by the discriminator to weight the value function estimates in the policy objective, leading to the following updates:
\begin{align}
    Q^{k+1} \leftarrow& \argmin_{Q} \E_{s, a, s' \sim \mathcal{D}}\left[ \left((R(s, a) +  \gamma \E_{a' \sim {\policy}^k(a'|s')}[Q^{k}(s', a')]) - Q_{target}(s, a)\right)^2 \right] \label{eq:dasco_q}\\ 
    \policy^{k+1} \leftarrow& \argmax_{\policy} \E_{s, a_{\mathcal{D}} \sim \mathcal{D}, a \sim \policy^k(a|s)}\left[ \dfrac{ D^k(s,a) }{ D^k( s,a_{\mathcal{D}}(s)) }Q^{k+1}(s, a) + \log D^k(s,a) \right],
    \label{eq:dasco_pi}
\end{align}
where $ a_{\mathcal{D}}(s) $ is the action from the offline dataset. 
The term $ D^k(s,a) $ down-weights the contribution of the gradient of the value function to the policy update if the discriminator deems the sampled action too unlikely. We further calibrate the probability $ D^k(s,a) $ by dividing it with the probability $ D^k(s,a_{\mathcal{D}}(s)) $ that the discriminator assigns to the dataset action $a_{\mathcal{D}}(s) $. It should be noted that during optimization the gradients are not propagated into these weights.

Next, we define the update rules for the auxiliary generator and the discriminator. We mix the samples from the $k^{th}$ iterate of the policy $\pi^k$ and the distribution $p_{aux}$ induced by the $k^{th}$ iterate of the auxiliary generator $G^k_{aux}$, that is, let $p_{mix} = \dfrac{ \pi^k + p_{aux} }{ 2 }$. At every iteration $k$, we update the $k^{th}$ iterate of the auxiliary generator $ G^{k}_{aux} $ and discriminator $D^k$ using the objectives:
\begin{align}
    G^{k+1}_{aux} \leftarrow& \argmin_{ G_{aux} } \mathbb{E}_{ x \sim p_{mix} } [ \log(1-D^k(s,a))] \label{eq:dasco_gaux}\\ 
    D^{k+1}  \leftarrow& \argmax_{D}  \mathbb{E}_{ x \sim p_{\mathcal{D}} } [\log(D^k(s,a))] + \mathbb{E}_{ x \sim p_{mix} } [ \log(1-D^k(s,a))]\label{eq:dasco_d}
\end{align}
\subsection{Algorithm summary}
\label{sec:algo_summary}

Algorithm \autoref{algo:dasco} provides a step-by-step description of our algorithm. At every training step, we sample a batch of transitions from the offline dataset and proceed to update the parameters of the value function, the policy, the auxiliary generator and the discriminator in that order.

\begin{algorithm}[H]
\begin{algorithmic}[1]
\STATE{Initialize Q-function $Q_\theta$, policy $\pi_\phi$, auxiliary generator $G_{aux, \psi}$, discriminator $D_\omega$}
\FOR{training step $k$ in \{1,\dots,N\}}
\STATE{$(s, a, r, s') \gets \mathcal{D}$: Sample a batch of transitions from the dataset}
\STATE{$\theta^{k+1} \gets$ Update Q-function $Q_\theta$ using the Bellman update in Eq.~\ref{eq:dasco_q}}
\STATE{$\phi^{k+1} \gets$ Update policy $\pi_\phi$ using the augmented objective in Eq.~\ref{eq:dasco_pi}}
\STATE{$\psi^{k+1} \gets$ Update auxiliary generator $G_{aux, \psi}$ using the objective in Eq.~\ref{eq:dasco_gaux}}
\STATE{$\omega^{k+1} \gets$ Update discriminator $D_\omega$ using mixed samples from $\pi_\phi$ and $G_{aux, \psi}$ as in Eq.~\ref{eq:dasco_d}}
\ENDFOR
\caption{\name{} algorithm summary}
\label{algo:dasco}
\end{algorithmic}
\end{algorithm}


\section{Experiments}
\label{sec:exp}
\renewcommand{\arraystretch}{1.2}


Our experiments aim at answering the following questions: 
\begin{enumerate}
    \item When learning from offline datasets that require combining actions from sub-optimal trajectories, does \name{} outperform existing methods that are based on distribution constraints? 
    \item On standard benchmarks such as D4RL~\cite{d4rl}, how does \name{} compare against recent methods? 
    \item Are both the dual generator and the probability ratio weight important for the performance of \name{}?
\end{enumerate}

\subsection{Comparisons on standard benchmarks and new datasets}

For our first set of experiments, we introduce four new datasets to simulate the challenges one might encounter when using offline RL algorithms on real world data. These datasets introduce additional learning challenges and require the algorithm to combine actions in different trajectories to obtain good performance. We use the existing AntMaze environments from the D4RL suite~\cite{d4rl}: antmaze-medium and antmaze-large. In these two environments, the algorithm controls an 8-DoF ``Ant" quadruped robot to navigate a 2D maze to reach desired goal locations. 
The D4RL benchmark generates the offline datasets for these two environments using two policies: 1. a low-level goal reaching policy that outputs torque commands to move the Ant to a nearby goal location and 2. a high-level waypoint generator to provide sub-goals that guide the low-level goal-reaching policy to the desired location.
We use the same two policies to generate two new classes of datasets. 

For the \texttt{noisy}
dataset, we add Gaussian noise to the action computed by the low-level goal-reaching policy. The noise variance depends on the 2D location of the Ant in the maze -- larger in some 2D regions than others. We intend this dataset to be representative of situations where the data generation policies are more deterministic in some states than others~\cite{kumar2022prefer} -- a robot picking up an object has many good options to approach the object, but when the robot grasps the object, its behavior should be more deterministic to ensure successful grasp without damaging or dropping the object \cite{10.5555/561828}.

For a \texttt{biased} dataset, in addition to adding Gaussian noise to the actions as it is done in the \texttt{noisy} dataset, we also add  bias to the actions computed by the low-level policy. The values of the bias also depend on the current 2D location of the Ant in the maze. This setting is meant to simulate learning from relabelled data, where the dataset was generated when the data generation policies were performing a different task than the tasks we are using the dataset to learn to perform. Relabelling offline data is a popular method for improving the performance of offline RL algorithms \cite{MBRL, Snell2022}, especially when we have much more data for some tasks than others \cite{mtopt}.
In the AntMaze environment, offline RL algorithms must combine data from sub-optimal trajectories to learn behaviors with high returns. In addition, \texttt{noisy} and \texttt{biased} datasets present a more challenging learning scenarios due to the added noise and systematic bias which vary non-uniformly based on the 2D location of the Ant.

\autoref{tab:hetero_antmaze} illustrates the performance comparison of our method and representative methods that enforce distribution constraints, either through optimizing a conservative lower bound of the value estimates (CQL) or only optimizing the policy on actions in the dataset using Advantage Weighted Regression \cite{peng2019awr} (IQL). Our method outperforms both CQL and IQL. In these tasks, to ensure a fair comparison between  different methods, we perform oracle offline policy selection to obtain the performance estimates for CQL, IQL, and our method. \edit{We describe how we tune hyper-parameters for the baselines in Appendix~\ref{sec:baseline_details}. We also include results of recent methods, such as EDAC \cite{EDAC} and BEAR \cite{bear}.} We also compare the performance on standard AntMaze tasks when learning from the datasets in the D4RL benchmark without modifications in \autoref{tab:d4rl_antmaze}. In these tasks, our method outperforms IQL by a large margin on two diverse datasets. 

\begin{table}[t]\centering
\caption{Performance comparison to baselines when learning from the \texttt{noisy} and \texttt{biased} AntMaze datasets. Our method outperforms the baselines significantly. The value in parenthesis indicates the standard deviation of mean episode return, computed over 3 different runs.}\label{tab:hetero_antmaze}
\small
\begin{tabular}{l||rrrr|r}
Dataset & \edit{BEAR} & \edit{EDAC} & CQL & IQL & \name{} (Ours) \\ \hline

antmaze-large-bias & - & - & 61.7 (3.5) & 41.0 (7.9) & 63.9 (6.0) \\

antmaze-large-noisy & - & - & 50.3 (2.3) & 39.0 (6.4) & 54.3 (2.0) \\

antmaze-medium-bias & 0.0 (0.0) & 0.0 (0.0) & 66.7 (2.9) & 48.0 (5.9) & 90.2 (2.4) \\

antmaze-medium-noisy & 0.0 (0.0) & 0.0 (0.0) & 55.7 (4.7) & 44.3 (1.7) & 86.3 (4.5) \\ \hline

\texttt{noisy} and \texttt{biased} antmaze-v2 total & - & -  & 234.4 & 172.3 &  \textbf{294.7} \\ \hline
\end{tabular}
\end{table}


\begin{table}[t]\centering
\caption{ \edit{Performance comparison to distribution-constrained baselines on AntMaze tasks in D4RL. Our algorithm outperforms the baselines when learning from the diverse and play datasets.} }\label{tab:d4rl_antmaze}
\small
\begin{tabular}{l | |rr | r}
Dataset & CQL & IQL & \name{} (Ours) \\ \hline
antmaze-umaze & 97.0 (0.8) & 90.3 (1.9) & 99.2 (0.0) \\
antmaze-umaze-diverse & 58.7 (12.2) & 70.3 (4.6) & 89.0 (1.7) \\
antmaze-medium-play & 77.0 (1.6) & 82.7 (0.5) & 92.3 (1.5) \\
antmaze-medium-diverse & 80.0 (0.0) & 82.3 (1.9) & 87.1 (0.4) \\
antmaze-large-play & 53.3 (4.6) & 55.7 (3.1) & 64.4 (1.7) \\
antmaze-large-diverse & 48.0 (2.9) & 50.0 (3.6) & 74.1 (2.8) \\
\hline
antmaze total & 414.0 & 431.3 & \textbf{506.1} \\
\hline
\end{tabular}
\vspace{-0.4cm}
\end{table}

\begin{table}[t]\centering
\caption{Performance comparison with recent offline RL algorithms on the Gym locomotion tasks}\label{tab:d4rl_locomotion}
\scriptsize
\setlength{\tabcolsep}{0.7em}
\hspace*{-1.65cm}
\begin{tabular}{l||rrrrrrrrr|r}
Dataset &BC &10\%BC & DT \cite{chen2021decision} & AWAC \cite{AWAC} & Onestep RL \cite{one_step_RL} & TD3+BC \cite{td3bc} & \edit{COMBO} \cite{yu2021combo} & CQL & IQL & \name{} (Ours) \\\hline
halfcheetah-medium-replay & 36.6 &40.6 &36.6 &40.5 &38.1 & 44.6 & 55.1 & 45.5 & 44.2 & 44.7  \\
hopper-medium-replay & 18.1 &75.9 &82.7 &37.2 & 97.5 & 60.9 & 89.5 & 95.0 & 94.7 & 101.7  \\
walker2d-medium-replay & 26.0 &62.5 &66.6 &27.0 &49.5 & 81.8 & 56.0 & 77.2 & 73.9 & 74.5 \\
halfcheetah-medium-expert & 55.2 & 92.9 &86.8 &42.8 & 93.4 & 90.7 & 90.0 & 91.6 & 86.7 & 93.8  \\
hopper-medium-expert & 52.5 & 110.9 & 107.6 &55.8 &103.3 &98.0 & 111.1 & 105.4 & 91.5 & 110.9 \\
walker2d-medium-expert & 107.5 & 109.0 & 108.1 & 74.5 & 113.0 & 110.1 & 103.3 & 108.8 & 109.6 & 109.3  \\ \hline
locomotion total & 295.9 & 491.8 & 488.4 & 277.8 & 494.8 & 486.1 & 505 & 523.5 & 500.6 & 534.9 \\
\hline
\end{tabular}
\end{table}

By comparing the results in \autoref{tab:hetero_antmaze} (learning from \texttt{noisy} and \texttt{biased} datasets) and \autoref{tab:d4rl_antmaze} (learning from existing offline datasets in D4RL), we also note that our proposed algorithm outperforms distribution-constraint offline RL algorithms (CQL, IQL) more consistently when tested on the \texttt{noisy} and \texttt{biased} datasets. For the results in these two tables, the definition of the antmaze-medium and antmaze-large environments are the same. The only axis of variation in the learning setup is the noise and systematic bias added to the actions of the dataset generation policies. We therefore conclude that our algorithm is more robust to the noise and systematic bias added to the actions than distribution-constrained offline RL algorithms.


Next, we evaluate our approach on Gym locomotion tasks from the standard D4RL suite. The performance results on these tasks are illustrated in \autoref{tab:d4rl_locomotion}. Our method is competitive with BC, one-step offline RL methods \cite{brandfonbrener2021offline}, and multi-step distribution-constraint RL  methods \cite{kostrikov2021offline, kumar2020conservative}. This is not surprising because in these tasks, the offline dataset contains a large number of trajectories with high returns. \edit{In \autoref{tab:d4rl_locomotion}, 10\%BC refers to performing BC using the top $10\%$ trajectories with the highest episode return. }





\subsection{Ablations}

We conduct three different sets of experiments to gain more insights into our algorithm. The first experiment measures the importance of having an auxiliary generator. We recall that there are two benefits to having the auxiliary generator. Firstly, without the auxiliary generator, the generator does not in general match the data distribution (\autoref{thm_disjoint}). As such, the discriminator has an unfair advantage in learning how to distinguish between real and generated examples. Secondly, the auxiliary generator plays the role of a support player and learns to output actions that are assigned non-zero probability by the data distribution, but have low Q values. The support player allows the policy to concentrate on in-support maximization of the Q-function (\autoref{thm_joint_in_support_max}). \autoref{tab:abs_aux_gen} demonstrates that having an auxiliary generator clearly leads to a performance improvement across different task families, from Gym locomotion tasks to AntMaze navigation tasks.

The second experiment compares the performance of the policy and the auxiliary generator on a subset of the Gym locomotion and AntMaze tasks (\autoref{tab:policy_vs_aux}). The difference in the performance of the policy and auxiliary generator illustrates their specialization of responsibility: the policy learns to output actions that lead to good performance, while the auxiliary generator learns to model the ``remainder" of the data distribution. If this ``remainder" also contains good action, then the auxiliary generator will have non-trivial performance. Otherwise, the auxiliary generator will have poor performance. 

In the Gym locomotion tasks, the auxiliary generator has non-trivial performance, but it is still worse than the policy. This demonstrates that: 1. By optimizing the policy to maximize the long-term return and the discriminator function, the policy can outperform the auxiliary generator, which only maximizes the discriminator function, 2. The dataset contains a large fraction of medium performance level actions contained in continuous trajectories, which the auxiliary generator has learnt to output. In contrast, in the \texttt{bias} and \texttt{noisy} AntMaze tasks, the auxiliary generator fails to obtain non-zero performance while the policy has strong performance. This reflects the necessity of carefully picking a subset of the in-support actions to obtain good performance.


\begin{table}[t]\centering
\caption{Ablation for training without and with auxiliary generator. The dual generator technique, which trains the auxiliary generator in addition to the policy, is crucial to obtain good performance.}\label{tab:abs_aux_gen}
\small
\begin{tabular}{l||rr}
Dataset & Without & With \\ \hline

halfcheetah-medium-expert & 79.8 (3.4) &  93.8 (0.1) \\
hopper-medium-expert & 95.1 (1.6) & 110.9 (0.8) \\ 

antmaze-large-bias & 55.0 (2.3) & 63.9 (6.0) \\

antmaze-large-noisy & 45.1 (1.8) & 54.3 (2.0) \\


\hline
\end{tabular}
\vspace{-0.2cm}
\end{table}

\begin{table}[!htp]\centering
\caption{Policy vs Auxiliary Generator. The auxiliary generator has reasonable performance on the easier locomotion tasks and is significantly worse than the policy on the harder AntMaze tasks.}\label{tab:policy_vs_aux}
\small
\begin{tabular}{l||rr}
Dataset & Auxiliary Generator & Policy \\ \hline

halfcheetah-medium-expert & 48.5 (2.1) & 93.8 (0.1) \\
hopper-medium-expert & 70.4 (0.9) & 110.9 (0.8) \\

antmaze-large-bias & 0.0 (0.0) & 63.9 (6.0) \\

antmaze-large-noisy  & 0.0 (0.0)  & 54.3 (2.0) \\

\hline
\end{tabular}
\end{table}

The third set of experiments illustrates the importance of weighing the value function in the policy objective by the probability computed by the discriminator, as described in Eq.~\ref{eq:dasco_pi}. Doing so provides a second layer of protection against exploitation of  errors in the value function by the policy. \autoref{tab:abl_dynamic_weighting} illustrates that this is very important for the AntMaze tasks, which require combining optimal and sub-optimal trajectories to obtain good performance. Perhaps this is because learning from such trajectories necessitates many rounds of offline policy evaluation and improvement steps, with each round creating an opportunity for the policy to exploit the errors in the value estimates. 
On the other hand, the dynamic weight is less important in the Gym locomotion tasks, presumably because a significant fraction of the corresponding offline datasets has high returns and therefore incorporating sub-optimal data is less criticial to obtain high performance.

\begin{table}[t]\centering
\caption{Ablation for dynamic weighting of value function estimates in the policy objective. When learning from datasets that require combining actions across trajectories, such as the AntMaze tasks, using the dynamic weighting is vital to obtaining good performance.}\label{tab:abl_dynamic_weighting}
\small
\begin{tabular}{l||rr}
Dataset & Without & With \\ \hline

halfcheetah-medium-expert & 91.1 (1.1) & 93.8 (0.1)  \\
hopper-medium-expert & 106.7 (2.9) & 110.9 (0.8) \\

antmaze-large-play & 0.0 (0.0) & 64.4 (1.7) \\ 
antmaze-large-diverse & 0.0 (0.0) & 74.1 (2.8) \\ 

\hline
\end{tabular}
\end{table}

\section{Conclusions}

In this paper, we introduced \name{}, a GAN-based offline RL method that addresses the challenges of training policies as generators with a discriminator to minimize deviation from the behavior policy by means of two modifications: an auxiliary generator to turn the GAN loss into a support constraint, and a value function weight in the policy objective. The auxiliary generator makes it possible for the policy to focus on maximizing the value function without needing to match the \emph{entirety} of the data distribution, only that part of it that has high value, effectively turning the standard distributional constraint that would be enforced by a conventional GAN into a kind of support constraint. This technique may in fact be of interest in other settings where there is a need to maximize some objective in addition to fooling a discriminator, and applications of this approach outside of reinforcement learning are an exciting direction for future work. Further, since our method enables GAN-based strategies to attain good results on a range of offline RL benchmark tasks, it would also be interesting in future work to consider other types of GAN losses that induce different divergence measures. We also plan to explore robust methods for offline policy and hyper-parameter selection in the future.

\edit{\textbf{Potential negative societal impact} Because our focus is on developing a generic offline RL algorithm and not its application in any particular domain, our method inherits the potential negative societal impacts that any offline RL algorithm might have, such as data bias and lack of robustness to distributional shift. We also note that how to perform effective offline policy selection and evaluation in the general case remain challenging. Therefore, the offline learned RL policies might not be easily deployed to solve real world problems where online policy evaluation is expensive.}

\bibliography{neurips_2022}{}

\begin{thebibliography}{51}
\providecommand{\natexlab}[1]{#1}
\providecommand{\url}[1]{\texttt{#1}}
\expandafter\ifx\csname urlstyle\endcsname\relax
  \providecommand{\doi}[1]{doi: #1}\else
  \providecommand{\doi}{doi: \begingroup \urlstyle{rm}\Url}\fi

\bibitem[Agarwal et~al.(2020)Agarwal, Schuurmans, and
  Norouzi]{agarwal2019optimistic}
Rishabh Agarwal, Dale Schuurmans, and Mohammad Norouzi.
\newblock An optimistic perspective on offline reinforcement learning.
\newblock In \emph{International Conference on Machine Learning (ICML)}, 2020.

\bibitem[An et~al.(2021)An, Moon, Kim, and Song]{EDAC}
Gaon An, Seungyong Moon, Jang{-}Hyun Kim, and Hyun~Oh Song.
\newblock Uncertainty-based offline reinforcement learning with diversified
  q-ensemble.
\newblock \emph{CoRR}, abs/2110.01548, 2021.
\newblock URL \url{https://arxiv.org/abs/2110.01548}.

\bibitem[Bai et~al.(2022)Bai, Wang, Yang, Deng, Garg, Liu, and Wang]{PBRL}
Chenjia Bai, Lingxiao Wang, Zhuoran Yang, Zhihong Deng, Animesh Garg, Peng Liu,
  and Zhaoran Wang.
\newblock Pessimistic bootstrapping for uncertainty-driven offline
  reinforcement learning, 2022.
\newblock URL \url{https://arxiv.org/abs/2202.11566}.

\bibitem[Boyd and Vandenberghe(2004)]{boyd2004convex}
Stephen Boyd and Lieven Vandenberghe.
\newblock \emph{Convex optimization}.
\newblock Cambridge university press, 2004.

\bibitem[Brandfonbrener et~al.(2021{\natexlab{a}})Brandfonbrener, Whitney,
  Ranganath, and Bruna]{brandfonbrener2021offline}
David Brandfonbrener, William~F Whitney, Rajesh Ranganath, and Joan Bruna.
\newblock Offline rl without off-policy evaluation.
\newblock \emph{arXiv preprint arXiv:2106.08909}, 2021{\natexlab{a}}.

\bibitem[Brandfonbrener et~al.(2021{\natexlab{b}})Brandfonbrener, Whitney,
  Ranganath, and Bruna]{one_step_RL}
David Brandfonbrener, William~F. Whitney, Rajesh Ranganath, and Joan Bruna.
\newblock Offline {RL} without off-policy evaluation.
\newblock \emph{CoRR}, abs/2106.08909, 2021{\natexlab{b}}.
\newblock URL \url{https://arxiv.org/abs/2106.08909}.

\bibitem[Chen et~al.(2021)Chen, Lu, Rajeswaran, Lee, Grover, Laskin, Abbeel,
  Srinivas, and Mordatch]{chen2021decision}
Lili Chen, Kevin Lu, Aravind Rajeswaran, Kimin Lee, Aditya Grover, Michael
  Laskin, Pieter Abbeel, Aravind Srinivas, and Igor Mordatch.
\newblock Decision transformer: Reinforcement learning via sequence modeling.
\newblock \emph{arXiv preprint arXiv:2106.01345}, 2021.

\bibitem[Chen et~al.(2022)Chen, Ghadirzadeh, Yu, Gao, Wang, Li, Liang, Finn,
  and Zhang]{chen2022latent}
Xi~Chen, Ali Ghadirzadeh, Tianhe Yu, Yuan Gao, Jianhao Wang, Wenzhe Li, Bin
  Liang, Chelsea Finn, and Chongjie Zhang.
\newblock Latent-variable advantage-weighted policy optimization for offline
  rl.
\newblock \emph{arXiv preprint arXiv:2203.08949}, 2022.

\bibitem[Chen et~al.(2019)Chen, Zhou, Wang, Wang, Wu, and Ross]{BAIL}
Xinyue Chen, Zijian Zhou, Zheng Wang, Che Wang, Yanqiu Wu, and Keith Ross.
\newblock Bail: Best-action imitation learning for batch deep reinforcement
  learning, 2019.
\newblock URL \url{https://arxiv.org/abs/1910.12179}.

\bibitem[Daoudi et~al.(2022)Daoudi, Barlier, Santos, and
  Virmaux]{daoudi2022density}
Paul Daoudi, Merwan Barlier, Ludovic~Dos Santos, and Aladin Virmaux.
\newblock Density estimation for conservative q-learning, 2022.
\newblock URL \url{https://openreview.net/forum?id=liV-Re74fK}.

\bibitem[Fu et~al.(2020)Fu, Kumar, Nachum, Tucker, and Levine]{d4rl}
J.~Fu, A.~Kumar, O.~Nachum, G.~Tucker, and S.~Levine.
\newblock D4rl: Datasets for deep data-driven reinforcement learning.
\newblock In \emph{arXiv}, 2020.
\newblock URL \url{https://arxiv.org/pdf/2004.07219}.

\bibitem[Fujimoto and Gu(2021{\natexlab{a}})]{fujimoto2021minimalist}
Scott Fujimoto and Shixiang~Shane Gu.
\newblock A minimalist approach to offline reinforcement learning.
\newblock \emph{arXiv preprint arXiv:2106.06860}, 2021{\natexlab{a}}.

\bibitem[Fujimoto and Gu(2021{\natexlab{b}})]{td3bc}
Scott Fujimoto and Shixiang~Shane Gu.
\newblock A minimalist approach to offline reinforcement learning.
\newblock \emph{CoRR}, abs/2106.06860, 2021{\natexlab{b}}.
\newblock URL \url{https://arxiv.org/abs/2106.06860}.

\bibitem[Fujimoto et~al.(2018)Fujimoto, Meger, and Precup]{bcq}
Scott Fujimoto, David Meger, and Doina Precup.
\newblock Off-policy deep reinforcement learning without exploration, 2018.
\newblock URL \url{https://arxiv.org/abs/1812.02900}.

\bibitem[Furuta et~al.(2022)Furuta, Matsuo, and Gu]{furuta2022generalized}
Hiroki Furuta, Yutaka Matsuo, and Shixiang~Shane Gu.
\newblock Generalized decision transformer for offline hindsight information
  matching.
\newblock In \emph{International Conference on Learning Representations}, 2022.
\newblock URL \url{https://openreview.net/forum?id=CAjxVodl_v}.

\bibitem[Goodfellow et~al.(2014{\natexlab{a}})Goodfellow, Pouget-Abadie, Mirza,
  Xu, Warde-Farley, Ozair, Courville, and Bengio]{gan}
I.~Goodfellow, J.~Pouget-Abadie, M.~Mirza, B.~Xu, D.~Warde-Farley, S.~Ozair,
  A.~Courville, and Y.~Bengio.
\newblock Generative adversarial nets.
\newblock In \emph{Neural Information Processing Systems (NIPS)},
  2014{\natexlab{a}}.

\bibitem[Goodfellow et~al.(2014{\natexlab{b}})Goodfellow, Pouget-Abadie, Mirza,
  Xu, Warde-Farley, Ozair, Courville, and Bengio]{GoodfellowPMXWOCB14}
Ian~J. Goodfellow, Jean Pouget-Abadie, Mehdi Mirza, Bing Xu, David
  Warde-Farley, Sherjil Ozair, Aaron Courville, and Yoshua Bengio.
\newblock Generative adversarial networks, 2014{\natexlab{b}}.
\newblock URL \url{https://arxiv.org/abs/1406.2661}.

\bibitem[Haarnoja et~al.(2018)Haarnoja, Zhou, Abbeel, and Levine]{sac}
T.~Haarnoja, A.~Zhou, P.~Abbeel, and S.~Levine.
\newblock Soft actor-critic: Off-policy maximum entropy deep reinforcement
  learning with a stochastic actor.
\newblock In \emph{arXiv}, 2018.
\newblock URL \url{https://arxiv.org/pdf/1801.01290.pdf}.

\bibitem[Hausman et~al.(2017)Hausman, Chebotar, Schaal, Sukhatme, and
  Lim]{intentiongan}
Karol Hausman, Yevgen Chebotar, Stefan Schaal, Gaurav~S. Sukhatme, and
  Joseph~J. Lim.
\newblock Multi-modal imitation learning from unstructured demonstrations using
  generative adversarial nets.
\newblock In Isabelle Guyon, Ulrike von Luxburg, Samy Bengio, Hanna~M. Wallach,
  Rob Fergus, S.~V.~N. Vishwanathan, and Roman Garnett, editors,
  \emph{NeurIPS}, pages 1235--1245, 2017.

\bibitem[Ho and Ermon(2016)]{gail}
J.~Ho and S.~Ermon.
\newblock Generative adversarial imitation learning.
\newblock In \emph{Neural Information Processing Systems (NIPS)}, 2016.

\bibitem[Jang et~al.(2022)Jang, Lee, and Kim]{jang2022gptcritic}
Youngsoo Jang, Jongmin Lee, and Kee-Eung Kim.
\newblock {GPT}-critic: Offline reinforcement learning for end-to-end
  task-oriented dialogue systems.
\newblock In \emph{International Conference on Learning Representations}, 2022.
\newblock URL \url{https://openreview.net/forum?id=qaxhBG1UUaS}.

\bibitem[Jaques et~al.(2019)Jaques, Ghandeharioun, Shen, Ferguson, Lapedriza,
  Jones, Gu, and Picard]{jaques2019way}
Natasha Jaques, Asma Ghandeharioun, Judy~Hanwen Shen, Craig Ferguson, Agata
  Lapedriza, Noah Jones, Shixiang Gu, and Rosalind Picard.
\newblock Way off-policy batch deep reinforcement learning of implicit human
  preferences in dialog.
\newblock \emph{arXiv preprint arXiv:1907.00456}, 2019.

\bibitem[Kalashnikov et~al.(2021)Kalashnikov, Varley, Chebotar, Swanson,
  Jonschkowski, Finn, Levine, and Hausman]{mtopt}
Dmitry Kalashnikov, Jacob Varley, Yevgen Chebotar, Benjamin Swanson, Rico
  Jonschkowski, Chelsea Finn, Sergey Levine, and Karol Hausman.
\newblock Mt-opt: Continuous multi-task robotic reinforcement learning at
  scale.
\newblock \emph{CoRR}, abs/2104.08212, 2021.
\newblock URL \url{https://arxiv.org/abs/2104.08212}.

\bibitem[Kingma and Ba(2014)]{adam}
Diederik~P. Kingma and Jimmy Ba.
\newblock Adam: A method for stochastic optimization, 2014.
\newblock URL \url{https://arxiv.org/abs/1412.6980}.

\bibitem[Kostrikov et~al.(2021{\natexlab{a}})Kostrikov, Nair, and
  Levine]{kostrikov2021offlineb}
Ilya Kostrikov, Ashvin Nair, and Sergey Levine.
\newblock Offline reinforcement learning with implicit q-learning.
\newblock \emph{arXiv preprint arXiv:2110.06169}, 2021{\natexlab{a}}.

\bibitem[Kostrikov et~al.(2021{\natexlab{b}})Kostrikov, Tompson, Fergus, and
  Nachum]{kostrikov2021offline}
Ilya Kostrikov, Jonathan Tompson, Rob Fergus, and Ofir Nachum.
\newblock Offline reinforcement learning with fisher divergence critic
  regularization.
\newblock \emph{arXiv preprint arXiv:2103.08050}, 2021{\natexlab{b}}.

\bibitem[Kumar(2019)]{kumar_blog}
Aviral Kumar.
\newblock Data-driven deep reinforcement learning.
\newblock \url{https://bair.berkeley.edu/blog/2019/12/05/bear/}, 2019.
\newblock {BAIR} Blog.

\bibitem[Kumar et~al.(2019{\natexlab{a}})Kumar, Fu, Soh, Tucker, and
  Levine]{bear}
Aviral Kumar, Justin Fu, Matthew Soh, George Tucker, and Sergey Levine.
\newblock Stabilizing off-policy q-learning via bootstrapping error reduction.
\newblock In \emph{Neural Information Processing Systems (NeurIPS)},
  2019{\natexlab{a}}.

\bibitem[Kumar et~al.(2019{\natexlab{b}})Kumar, Fu, Soh, Tucker, and
  Levine]{kumar2019stabilizing}
Aviral Kumar, Justin Fu, Matthew Soh, George Tucker, and Sergey Levine.
\newblock Stabilizing off-policy q-learning via bootstrapping error reduction.
\newblock In \emph{Advances in Neural Information Processing Systems}, pages
  11761--11771, 2019{\natexlab{b}}.

\bibitem[Kumar et~al.(2020)Kumar, Zhou, Tucker, and
  Levine]{kumar2020conservative}
Aviral Kumar, Aurick Zhou, George Tucker, and Sergey Levine.
\newblock Conservative q-learning for offline reinforcement learning.
\newblock \emph{arXiv preprint arXiv:2006.04779}, 2020.

\bibitem[Kumar et~al.(2022)Kumar, Hong, Singh, and Levine]{kumar2022prefer}
Aviral Kumar, Joey Hong, Anikait Singh, and Sergey Levine.
\newblock When should we prefer offline reinforcement learning over behavioral
  cloning?, 2022.

\bibitem[Levine et~al.(2020)Levine, Kumar, Tucker, and Fu]{levine2020offline}
Sergey Levine, Aviral Kumar, George Tucker, and Justin Fu.
\newblock Offline reinforcement learning: Tutorial, review, and perspectives on
  open problems.
\newblock \emph{arXiv preprint arXiv:2005.01643}, 2020.

\bibitem[Li et~al.(2017)Li, Song, and Ermon]{infogail}
Yunzhu Li, Jiaming Song, and Stefano Ermon.
\newblock Infogail: Interpretable imitation learning from visual
  demonstrations.
\newblock In Isabelle Guyon, Ulrike von Luxburg, Samy Bengio, Hanna~M. Wallach,
  Rob Fergus, S.~V.~N. Vishwanathan, and Roman Garnett, editors,
  \emph{NeurIPS}, pages 3812--3822, 2017.

\bibitem[Liu et~al.(2022)Liu, Tang, Li, and Luo]{liu2022robust}
Liu Liu, Ziyang Tang, Lanqing Li, and Dijun Luo.
\newblock Robust imitation learning from corrupted demonstrations, 2022.
\newblock URL \url{https://openreview.net/forum?id=UECzHrGio7i}.

\bibitem[Liu et~al.(2021)Liu, Zhang, Fu, Yang, and Wang]{zhihanliu}
Zhihan Liu, Yufeng Zhang, Zuyue Fu, Zhuoran Yang, and Zhaoran Wang.
\newblock Provably efficient generative adversarial imitation learning for
  online and offline setting with linear function approximation.
\newblock \emph{CoRR}, abs/2108.08765, 2021.
\newblock URL \url{https://arxiv.org/abs/2108.08765}.

\bibitem[Mao et~al.(2016)Mao, Li, Xie, Lau, Wang, and Smolley]{lsgan}
Xudong Mao, Qing Li, Haoran Xie, Raymond Y.~K. Lau, Zhen Wang, and Stephen~Paul
  Smolley.
\newblock Least squares generative adversarial networks, 2016.
\newblock URL \url{https://arxiv.org/abs/1611.04076}.

\bibitem[Meng et~al.(2022)Meng, Wen, Yang, chenyang le, yun Li, Zhang, Wen,
  Zhang, Wang, and XU]{meng2022offline}
Linghui Meng, Muning Wen, Yaodong Yang, chenyang le, Xi~yun Li, Haifeng Zhang,
  Ying Wen, Weinan Zhang, Jun Wang, and Bo~XU.
\newblock Offline pre-trained multi-agent decision transformer, 2022.
\newblock URL \url{https://openreview.net/forum?id=W08IqLMlMer}.

\bibitem[Murray et~al.(1994)Murray, Sastry, and Zexiang]{10.5555/561828}
Richard~M. Murray, S.~Shankar Sastry, and Li~Zexiang.
\newblock \emph{A Mathematical Introduction to Robotic Manipulation}.
\newblock CRC Press, Inc., USA, 1st edition, 1994.
\newblock ISBN 0849379814.

\bibitem[Nair et~al.(2020)Nair, Dalal, Gupta, and Levine]{AWAC}
Ashvin Nair, Murtaza Dalal, Abhishek Gupta, and Sergey Levine.
\newblock Accelerating online reinforcement learning with offline datasets.
\newblock \emph{CoRR}, abs/2006.09359, 2020.
\newblock URL \url{https://arxiv.org/abs/2006.09359}.

\bibitem[Paszke et~al.(2019)Paszke, Gross, Massa, Lerer, Bradbury, Chanan,
  Killeen, Lin, Gimelshein, Antiga, Desmaison, Köpf, Yang, DeVito, Raison,
  Tejani, Chilamkurthy, Steiner, Fang, Bai, and Chintala]{paszke2019pytorch}
Adam Paszke, Sam Gross, Francisco Massa, Adam Lerer, James Bradbury, Gregory
  Chanan, Trevor Killeen, Zeming Lin, Natalia Gimelshein, Luca Antiga, Alban
  Desmaison, Andreas Köpf, Edward Yang, Zach DeVito, Martin Raison, Alykhan
  Tejani, Sasank Chilamkurthy, Benoit Steiner, Lu~Fang, Junjie Bai, and Soumith
  Chintala.
\newblock Pytorch: An imperative style, high-performance deep learning library,
  2019.
\newblock URL \url{http://arxiv.org/abs/1912.01703}.
\newblock cite arxiv:1912.01703Comment: 12 pages, 3 figures, NeurIPS 2019.

\bibitem[Peng et~al.(2019)Peng, Kumar, Zhang, and Levine]{peng2019awr}
Xue~Bin Peng, Aviral Kumar, Grace Zhang, and Sergey Levine.
\newblock Advantage-weighted regression: Simple and scalable off-policy
  reinforcement learning.
\newblock \emph{arXiv preprint arXiv:1910.00177}, 2019.

\bibitem[Siegel et~al.(2020)Siegel, Springenberg, Berkenkamp, Abdolmaleki,
  Neunert, Lampe, Hafner, and Riedmiller]{siegel2020keep}
Noah~Y Siegel, Jost~Tobias Springenberg, Felix Berkenkamp, Abbas Abdolmaleki,
  Michael Neunert, Thomas Lampe, Roland Hafner, and Martin Riedmiller.
\newblock Keep doing what worked: Behavioral modelling priors for offline
  reinforcement learning.
\newblock \emph{arXiv preprint arXiv:2002.08396}, 2020.

\bibitem[Snell et~al.(2022)Snell, Yang, Fu, Su, and Levine]{Snell2022}
Charlie Snell, Mengjiao Yang, Justin Fu, Yi~Su, and Sergey Levine.
\newblock Context-aware language modeling for goal-oriented dialogue systems,
  2022.
\newblock URL \url{https://arxiv.org/abs/2204.10198}.

\bibitem[Sutton and Barto(2018)]{sutton1998}
Richard~S. Sutton and Andrew~G. Barto.
\newblock \emph{Reinforcement Learning: An Introduction}.
\newblock A Bradford Book, Cambridge, MA, USA, 2018.
\newblock ISBN 0262039249.

\bibitem[Sønderby et~al.(2016)Sønderby, Caballero, Theis, Shi, and
  Huszár]{gan_instance_noise}
Casper~Kaae Sønderby, Jose Caballero, Lucas Theis, Wenzhe Shi, and Ferenc
  Huszár.
\newblock Amortised map inference for image super-resolution, 2016.
\newblock URL \url{https://arxiv.org/abs/1610.04490}.

\bibitem[Vuong et~al.(2019)Vuong, Liu, Liu, Ciosek, Su, and Christensen]{MBRL}
Quan Vuong, Shuang Liu, Minghua Liu, Kamil Ciosek, Hao Su, and Henrik~Iskov
  Christensen.
\newblock Multi-task batch reinforcement learning with metric learning.
\newblock \emph{CoRR}, abs/1909.11373, 2019.
\newblock URL \url{http://arxiv.org/abs/1909.11373}.

\bibitem[Wang et~al.(2020)Wang, Novikov, {\.Z}o{\l}na, Springenberg, Reed,
  Shahriari, Siegel, Merel, Gulcehre, Heess, et~al.]{wang2020critic}
Ziyu Wang, Alexander Novikov, Konrad {\.Z}o{\l}na, Jost~Tobias Springenberg,
  Scott Reed, Bobak Shahriari, Noah Siegel, Josh Merel, Caglar Gulcehre,
  Nicolas Heess, et~al.
\newblock Critic regularized regression.
\newblock \emph{arXiv preprint arXiv:2006.15134}, 2020.

\bibitem[Watkins and Dayan(1992)]{watkins1992q}
Christopher~JCH Watkins and Peter Dayan.
\newblock Q-learning.
\newblock \emph{Machine learning}, 8\penalty0 (3-4):\penalty0 279--292, 1992.

\bibitem[Wu et~al.(2019)Wu, Tucker, and Nachum]{brac}
Yifan Wu, George Tucker, and Ofir Nachum.
\newblock Behavior regularized offline reinforcement learning.
\newblock \emph{arXiv preprint arXiv:1911.11361}, 2019.

\bibitem[Yu et~al.(2021)Yu, Kumar, Rafailov, Rajeswaran, Levine, and
  Finn]{yu2021combo}
Tianhe Yu, Aviral Kumar, Rafael Rafailov, Aravind Rajeswaran, Sergey Levine,
  and Chelsea Finn.
\newblock Combo: Conservative offline model-based policy optimization.
\newblock \emph{arXiv preprint arXiv:2102.08363}, 2021.

\bibitem[Zhou et~al.(2020)Zhou, Bajracharya, and Held]{zhou2020plas}
Wenxuan Zhou, Sujay Bajracharya, and David Held.
\newblock Plas: Latent action space for offline reinforcement learning.
\newblock \emph{arXiv preprint arXiv:2011.07213}, 2020.

\end{thebibliography}
\bibliographystyle{plainnat}

\section*{Checklist}

The checklist follows the references.  Please
read the checklist guidelines carefully for information on how to answer these
questions.  For each question, change the default \answerTODO{} to \answerYes{},
\answerNo{}, or \answerNA{}.  You are strongly encouraged to include a {\bf
justification to your answer}, either by referencing the appropriate section of
your paper or providing a brief inline description.  For example:
\begin{itemize}
  \item Did you include the license to the code and datasets? \answerYes{See Section~\ref{gen_inst}.}
  \item Did you include the license to the code and datasets? \answerNo{The code and the data are proprietary.}
  \item Did you include the license to the code and datasets? \answerNA{}
\end{itemize}

\begin{enumerate}

\item For all authors...
\begin{enumerate}
  \item Do the main claims made in the abstract and introduction accurately reflect the paper's contributions and scope?
    \answerYes{}
  \item Did you describe the limitations of your work?
    \answerYes{Please see the experimental results in Section 5 and avenues for future work in the Conclusion}
  \item Did you discuss any potential negative societal impacts of your work?
    \answerYes{We discussed the potential negative impacts in the Conclusion of the paper}
  \item Have you read the ethics review guidelines and ensured that your paper conforms to them?
    \answerYes{}
\end{enumerate}

\item If you are including theoretical results...
\begin{enumerate}
  \item Did you state the full set of assumptions of all theoretical results?
    \answerYes{We will include the proofs for all theorems in the Appendix}
        \item Did you include complete proofs of all theoretical results?
    \answerYes{We will include the proofs for all theorems in the Appendix}
\end{enumerate}

\item If you ran experiments...
\begin{enumerate}
  \item Did you include the code, data, and instructions needed to reproduce the main experimental results (either in the supplemental material or as a URL)?
    \answerYes{We will include these results in the Appendix}
  \item Did you specify all the training details (e.g., data splits, hyperparameters, how they were chosen)?
    \answerYes{We will include this information in the Appendix}
        \item Did you report error bars (e.g., with respect to the random seed after running experiments multiple times)?
    \answerYes{We will include this information in the Appendix}
        \item Did you include the total amount of compute and the type of resources used (e.g., type of GPUs, internal cluster, or cloud provider)?
    \answerYes{Please see Appendix~\ref{sec:additional_exp_details}}
\end{enumerate}

\item If you are using existing assets (e.g., code, data, models) or curating/releasing new assets...
\begin{enumerate}
  \item If your work uses existing assets, did you cite the creators?
    \answerNA{}
  \item Did you mention the license of the assets?
    \answerNA{}
  \item Did you include any new assets either in the supplemental material or as a URL?
    \answerNA{}
  \item Did you discuss whether and how consent was obtained from people whose data you're using/curating?
    \answerNA{}
  \item Did you discuss whether the data you are using/curating contains personally identifiable information or offensive content?
    \answerNA{}
\end{enumerate}

\item If you used crowdsourcing or conducted research with human subjects...
\begin{enumerate}
  \item Did you include the full text of instructions given to participants and screenshots, if applicable?
    \answerNA{}
  \item Did you describe any potential participant risks, with links to Institutional Review Board (IRB) approvals, if applicable?
    \answerNA{}
  \item Did you include the estimated hourly wage paid to participants and the total amount spent on participant compensation?
    \answerNA{}
\end{enumerate}

\end{enumerate}


\newpage
\appendix

\part*{Appendices}

\section{Proofs for theorems in Section~\ref{sec:dual_gen}}

\subsection{Proof for Theorem~\ref{thm_disjoint} }

In the following proof, we use $p_{\text{data}}$ to refer to the real data distribution, instead of $p_\mathcal{D}$ as in Section~\ref{sec:dual_gen}, to avoid confusion with the discriminator distribution.

We recall Theorem~\ref{thm_disjoint}:

\thmdisjoint*

The optimization problem in Eq.~\ref{eq:joint_nodual} is:
\begin{align*}
    \min_{G} \max_{D}  V(G, D) = \mathbb{E}_{ x \sim p_{\text{data}}} [\log(D(x))] + \mathbb{E}_{z \sim p(z)} [\log(1-D(G(z)))]  + \mathbb{E}_{z \sim p(z)} [ f ( G(z) ) ]
\end{align*}

The proof proceeds as follows: We first simplify the objective function into two terms. The first term is the Jensen–Shannon divergence between the data distribution and the distribution induced by the generator~\cite{gan}. The second term is the expected value of the secondary objective function $f$. We then show that the problem is convex, where strong duality holds. We then use the KKT conditions to find the functional form of the optimal solution, which gives us Theorem~\ref{thm_disjoint}.

We only prove the statement for discrete sample space, and we let $n$ be the size of the sample space -- the random variable $x$ can take on $n$ different values.

\textit{Proof}. Since the third term in the objective function is not a function of the discriminator $D$, for $G$ fixed, the optimal discriminator of Eq.~\ref{eq:joint_nodual} is $D^*_G(x) = \dfrac{ p_{\text{data}} (x) }{ p_{ \text{data} } (x) + p_{g }(x) }$ where $p_g$ is the distribution induced by the generator $G$.
 (similar to Prop 1 in \cite{gan} ).

Similarly to how \cite{gan} shows that the GAN objective in Eq.~\ref{eq:vanilla_gan} minimizes the JS divergence between the data distribution and the distribution induced by the generator, we can now rewrite the objective in Eq.~\ref{eq:joint_nodual} as:
\begin{align}
     & V(G, D^*_G)\\
     & = \mathbb{E}_{ x \sim p_{\text{data}}} [\log(D^*_G(x))] + \mathbb{E}_{z \sim p(z)} [\log(1-D^*_G(G(z)))]  + \mathbb{E}_{z \sim p(z)} [ f ( G(z) ) ] \\
     & = 2JSD( p_{\text{data}} || p_g ) + \mathbb{E}_{x \sim p_g} [ f ( x ) ] - \log 4 
\end{align}

For conciseness, let $g^{(i)} = p_g(x_i)$ be the probability that $p_g$ assigns to $x_i$  and $g = [g^{(1)}, \ldots, g^{(n)}]^T$ be a column vector containing the probabilities that $p_g$ assigns to each possible values of $x$, from $x_1$ to $x_n$. 

Similarly, let $f^{(i)} = f(x_i)$ be the value that the secondary objective $f$ assigns to $x_i$. We also overload the notation to let $f = [f^{(1)}, \ldots, f^{(n)}]^T$ be a column vector containing the values that the secondary objective $f$ assigns to each possible value of the random variable $x$, from $x_1$ to $x_n$. 

Also let $ p_{\text{data} }^{(i)} = p_{\text{data} } (x_i) $ be the probability that the data distribution assigns to $x_i$.

We can then rewrite the problem in Eq.~\ref{eq:joint_nodual} in a standard form \cite{boyd2004convex} as:
\begin{align}
\min_{g} \quad & 2JSD( p_{\text{data}} || p_g )  + g^T f \\
\textrm{s.t.} \quad & - g^{(i)} \leq 0 \label{eq:const_non_neg} \\
  & \mathbf{1}^T g - 1 = 0 \label{eq:const_normalized}
\end{align}
where $\mathbf{1}$ is a column vector of $1$, which has the same number of entries as the vector $g$. The constraint \ref{eq:const_non_neg} ensures that the probability that $p_g$ assigns to any $x$ is non-negative and the constraint \ref{eq:const_normalized} ensures the probabilities sum up to $1$.

The problem is convex because the objective function is a nonnegative weighted sum of two convex functions (JSD is convex because JSD is itself a nonnegative weighted sum of KL, which is a convex function). 

Strong duality also holds because Slater's condition holds. A strictly feasible point for Slater's condition to hold is the uniform distribution, i.e. $g^{(i)} = \dfrac{1}{n}, \forall i$.

The Lagrangian is:
\begin{align}
    L = 2JSD( p_{\text{data}} || p_g )  + g^T f - \sum_i \lambda^{(i)} g^{(i)} + \nu ( \mathbf{1}^T g - 1 )
\end{align}
where $\lambda^{(i)}$ and $\nu$ are the Lagrangian multipliers.

For any $i \in [1, n]$, the partial derivative of the Lagrangian with respect to $g^{(i)}$ is:
\begin{align}
    \pdv{L}{ g^{(i)} } & = log \left( \dfrac{ 2g^{(i)} } { p_{\text{data} }^{(i)} + g^{(i)} } \right) + f^{ (i) } - \lambda^{(i)} + \nu
 \end{align}

Let $g_*$ and $(\lambda_*, \nu_*)$ be the primal and dual optimal solutions of the optimization problem. As the strong duality holds, the variables $g_*$ and $(\lambda_*, \nu_*)$ must satisfy the KKT conditions. For any $i \in [1, n]$, the following holds:
\begin{align}
    -g_*^{(i)} & \leq 0 \\ 
    \mathbf{1}^T g_* - 1 & = 0 \\ 
    \lambda_*^{(i)} & \geq 0 \\ 
    \lambda_*^{(i)} g_*^{(i)} & = 0 \label{eq:kkt_com_slack} \\ 
    \pdv{L}{ g^{(i)} } = log \left( \dfrac{ 2g_*^{(i)} } { p_{\text{data} }^{(i)} + g_*^{(i)} } \right) + f^{ (i) } - \lambda_*^{(i)} + \nu_* & = 0 \label{eq:kkt_lang}
\end{align}

From \autoref{eq:kkt_lang}, we have $ \lambda_*^{(i)} = log \left( \dfrac{ 2g_*^{(i)} } { p_{\text{data} }^{(i)} + g_*^{(i)} } \right) + f^{ (i) } + \nu_* $, and substitute into \autoref{eq:kkt_com_slack}:

\begin{align}
    \left[ log \left( \dfrac{ 2g_*^{(i)} } { p_{\text{data} }^{(i)} + g_*^{(i)} } \right) + f^{ (i) } + \nu_* \right] g^*_i & = 0
\end{align}

We consider what happens when $g^*_i > 0$, due to complementary slackness, we have: 

\begin{align}
    \log \left( \dfrac{ 2g_*^{(i)} } { p_{\text{data} }^{(i)} + g_*^{(i)} } \right) + f^{ (i) } + \nu_* & = 0 \\ 
     \Longrightarrow g_*^{(i)} & = \dfrac{ p_{\text{data} }^{(i)}e^{-  f^{ (i) } - \nu_* } }{ ( 2 - e^{-  f^{ (i) } - \nu_* } ) } \\
     p^*_g(x_i) & = p_{\text{data} } (x_i) \dfrac{ e^{- f (x_i)  - \nu_* } } { 2 - e^{-  f(x_i)  - \nu_* } }
\end{align}
We can then pick an appropriate value for the Lagrange multiplier $\nu$ such that the probabilities $p^*_g(x_i)$ normalize to 1. QED.

\subsection{Proof for Theorem~\ref{thm_joint_in_support_max} }

In the following proof, we use $p_{\text{data}}$ to refer to the real data distribution, instead of $p_\mathcal{D}$ as in Section~\ref{sec:dual_gen}, to avoid confusion with the discriminator distribution.

Recall that we define $p_{mix}$ as $p_{mix} = \dfrac{ p_g + p_{aux} }{ 2 }$. Theorem \ref{thm_joint_in_support_max} is stated in reference to the optimization problem in Eq.~\ref{eq:joint_gan}, which we restate here:

\begin{align}
    \min_{G, G_{aux} } \max_{D} \quad & V(G, G_{aux}, D) = \mathbb{E}_{ x \sim p_{\text{data}} } [\log(D(x))] + \mathbb{E}_{ x \sim p_{mix} } [ \log(1-D(x))]  + \mathbb{E}_{ x \sim p_g } [ f ( x ) ] \label{eq:joint_gan_explicit_constraint_obj} 
\end{align}

where the first two terms in the objective function are the GAN objective and the last term is the secondary objective function. 

Similar to the proof for Theorem \ref{thm_disjoint}, we can rewrite the objective function in Eq.~\ref{eq:joint_gan_explicit_constraint_obj} as \cite{gan}:
\begin{align}
     & V(G, G_{aux}, D^*)\\
     & = 2JSD( p_{\text{data}} || \dfrac{ p_g + p_{aux} }{ 2 } ) + \mathbb{E}_{x \sim p_g} [ f ( x ) ] - \log 4 \label{eq:joint_gan_wih_jsd}
\end{align}

We are only interested in optimizing for the secondary objective function $f$ in the space of optimal GAN solutions. We therefore enforce that $p_{mix} = \dfrac{ p_g + p_{aux} }{ 2 } = p_{\text{data}}$, which makes the JSD term vanish in Eq.~\ref{eq:joint_gan_wih_jsd} and allows us to solve the following optimization problem.
\begin{align}
    \min_{G } \quad & \mathbb{E}_{x \sim p_g} [ f ( x ) ] \\ 
\textrm{s.t.} \quad & p_g \leq 2 p_{\text{data}} \label{eq:joint_gan_explicit_constraint_simplified_constraint_1} \\
& p_{aux} = 2 p_{\text{data}} - p_g \label{eq:joint_gan_explicit_constraint_simplified_constraint_2}
\end{align}


We claim that the solution to the optimization problem above is as follows. We define $x_0$ to be the element inside the support of the data distribution $p_{\text{data}}$ that minimizes $f$, i.e. $ x_0 = \underset{x \in \text{Supp}( p_{\text{data}} ) }{\arg\min} f(x) $.
The optimal primary generator $p^*_g$ assigns the following probability to $x_0$:
\begin{align}
    p^*_{g} (x_0) & = \begin{cases}
    2p_{\text{data}} (x_0) \quad  \text{if} \quad 2p_{\text{data}} (x_0) < 1 \\
    1 \quad \quad  \quad \quad \text{otherwise}
    \end{cases}
\end{align}

If the global maximum $x_0$ is not taking the full probability mass, the rest of the probability mass is redistributed to the next best in-support maxima, which 
we can define recursively:
\begin{align}
         \text{For} \,\, x_i \in \underset{x \in \text{Supp}( p_{\text{data}} )  \setminus \{ x_j \}_{j=0}^{i-1} }{\arg\min} f(x), \,\,
    p^*_{g} (x_i) & = \begin{cases}
      2p_{\text{data}} (x_i)\quad \quad \quad \quad \,\,\, \text{if} \quad \sum_{j=0}^i p^*_g(x_j) < 1 \\
      1 -  \sum_{j=0}^{i-1} p^*_g(x_j) \quad \,\, \text{if} \quad \sum_{j=0}^i p^*_g(x_j) > 1 \\ 
      0 \quad \quad \quad \quad \quad \quad \quad \quad  \text{if} \quad \sum_{j=0}^{i-1} p^*_g(x_j) = 1 
    \end{cases}       
\end{align}

\textit{Proof.}

We show the proof by contradiction. That is, assume that there exists another distribution $p^a_g$ with the following properties:
\begin{itemize}
    \item There exists $x$ where $p^a_g(x) \neq p_g^*(x)$
    \item $p^a_g$ satisfies the constraint (\ref{eq:joint_gan_explicit_constraint_simplified_constraint_1})-(\ref{eq:joint_gan_explicit_constraint_simplified_constraint_2})
    \item The value of the objective function achieved by $p^a_g$ is better than the value achieved by $p_g^*$. That is, $\mathbb{E}_{x \sim p^a_g} [ f ( x ) ] < \mathbb{E}_{x \sim p^*_g} [ f ( x ) ]$.
\end{itemize}
We will show that the existence of such a distribution $p^a_g$ will lead to contradiction,

We separate the analyses into three different cases, depending on the property of $p_g^*$: 
\begin{itemize}
    \item Case 1: $p_g^*$ assigns all probability mass to $x_0$
    \item Case 2: If $p_g^*$ assigns non-zero probability to $x$, then $p_g^* = 2p_{\text{data}} (x)$
    \item Case 3: There exists an $x$ where $ 2 p_{\text{data}} (x) >  p_g^*(x) > 0$ 
\end{itemize}

We will walk through the three cases independently and show the contradiction in each case.

\textbf{Case 1:} $p^*_{g}$ assigns the full probability mass to $x_0$, that is $p^*_{g} (x_0) = 1$, and assigns zero probability to every $x$ not equal to $x_0$. Without loss of generality, we consider $p_g$ that assigns non-zero probability to a $x_k \neq x_0$, assigns the remaining probability mass to $x_0$, and assigns zero probability to all $x$ that is not equal to either $x_0$ or $x_k$. That is, assume there exists $p^a_g$ such that:
\begin{align}
    0 > p^a_g (x_0) & > 1 \\ 
    p^a_g (x_k) & = 1 - p^a_g (x_0) > 0 \text{ for some } x_k \in \text{Supp}( p_{\text{data}} ) \\ 
    \mathbb{E}_{ x \sim p^*_g } [ f ( x ) ] - \mathbb{E}_{ x \sim p^a_g } [ f ( x ) ]& > 0 \label{eq:case_1_contra}
\end{align}
where $x_k \in \text{Supp}( p_{\text{data}} ) $ follows from constraint \ref{eq:joint_gan_explicit_constraint_simplified_constraint_1} ($p_g \leq 2 p_{\text{data}} $, and thus $p^a_g$ can only assign non-zero probability to $x$ within the support of $p_{\text{data}}$).
We can then show that:
\begin{align}
\mathbb{E}_{ x \sim p^*_g } & [ f ( x ) ] - \mathbb{E}_{ x \sim p^a_g } [ f ( x ) ] \\
= & f(x_0) - p^a_g (x_0) f(x_0) - p^a_g (x_k) f(x_k) \\
= & ( 1 - p^a_g (x_0) ) f(x_0) - p^a_g (x_k) f(x_k) \\
= & p^a_g (x_k) f(x_0) - p^a_g (x_k) f(x_k) \\
= & p^a_g (x_k) [ f(x_0) - f(x_k) ] \leq 0 \text{ (contradiction with Eq.} \ref{eq:case_1_contra} )
\end{align}
where the last inequity follows from these two facts:
\begin{align}
    x_0 & = \underset{x \in \text{Supp}( p_{\text{data}} ) }{\arg\min} f(x) \\
    x_k & \in \text{Supp}( p_{\text{data}} )
\end{align}

\textbf{Case 2:} 
\begin{align}
    p^*_{g} (x) & = \begin{cases}
    2p_{\text{data}} (x) \quad  \text{if} \quad p^*_{g} (x) > 0 \\
    0 \quad \quad  \quad \quad \text{otherwise}
    \end{cases}
\end{align}

Let $\{ x_0, \ldots, x_i \}$ be the set of x where $p^*_{g} (x) > 0$, then we also require that $\sum_{j=0}^i p^*_{g} (x) = 1 $.

Without loss of generality, we assume a distribution $p^a_g$ exists with the following properties. There exists $x_m, x_n$ such that:
\begin{align}
    p^*_{g} (x_m) = 2p_{\text{data}} (x_m) > 0 \quad & \text{ and } \quad  p^a_g (x_m) < 2p_{\text{data}} (x_m) \\
    p^*_{g} (x_n) = 0 \quad & \text{ and } \quad p^a_g (x_n) = 2p_{\text{data}} (x_m) - p^a_g (x_m) > 0  \\
    p^*_{g} (x) & = p^a_g (x) \text{ otherwise (that is, for all } x \notin \{x_m, x_n\}) \\ 
    \mathbb{E}_{ x \sim p^*_g } [ f ( x ) ] - \mathbb{E}_{ x \sim p^a_g } [ f ( x ) ] & > 0 \label{eq:case_2_contra}
\end{align}
We note that $f(x_m) \leq f(x_n)$ since $p^*_{g}$ assigns non-zero probability to $x_m$ and assigns zero probability to $x_n$.

We can show that:
\begin{align}
\mathbb{E}_{ x \sim p^*_g } & [ f ( x ) ] - \mathbb{E}_{ x \sim p^a_g } [ f ( x ) ] \\
= & p^*_g (x_m) f(x_m) - p^a_g (x_m) f(x_m) - p^a_g (x_n) f(x_n) \\ 
= & p^*_g (x_m) f(x_m) - p^a_g (x_m) f(x_m) - p^a_g (x_n) f(x_n) \\ 
= & p^*_g (x_m) f(x_m) - p^a_g (x_m) f(x_m) - (2p_{\text{data}} (x_m) - p^a_g (x_m)) f(x_n) \\ 
= & p^*_g (x_m) f(x_m) - p^a_g (x_m) f(x_m) - 2p_{\text{data}} (x_m) f(x_n) + p^a_g (x_m) f(x_n) \\ 
= & p^*_g (x_m) f(x_m) - p^a_g (x_m) f(x_m) - p^*_g (x_m) f(x_n) + p^a_g (x_m) f(x_n) \\ 
= & p^*_g (x_m) [f(x_m) - f(x_n)] - p^a_g (x_m) [f(x_m) - f(x_n)] \\ 
= & [f(x_m) - f(x_n)] [p^*_g (x_m)  - p^a_g (x_m)] \leq 0 \text{ (contradiction with Eq.} \ref{eq:case_2_contra} )
\end{align}
where the last inequality is true because $f(x_m) \leq f(x_n)$ as we noted above, and $p^*_g (x_m) = 2p_{\text{data}} (x_m) > p^a_g (x_m)$.

\textbf{Case 3:}

There exists $x_i$ such that $2p_{\text{data}} (x_i) > p^*_g(x_i) > 0$.
For all $x \neq x_i$:
\begin{align}
    p^*_{g} (x) & = \begin{cases}
    2p_{\text{data}} (x) \quad  \text{if} \quad p^*_{g} (x) > 0 \\
    0 \quad \quad  \quad \quad \text{otherwise}
    \end{cases}
\end{align}

Let $\{ x_0, \ldots, x_i \}$ be the set of x where $p^*_{g} (x) > 0$, we also require $\sum_{j=0}^i p^*_{g} (x) = 1 $.

Without loss of generality, there are three cases we need to consider for the distribution $p^a_g$, each yielding a contradiction:
\begin{itemize}
    \item $p^a_g (x_i) = p^*_g(x_i)$, but there exists $x$ such that $p^a_g (x) \neq p^*_g(x)$.
    \item $p^a_g (x_i) > p^*_g(x_i)$.
    \item $p^a_g (x_i) < p^*_g(x_i)$.
\end{itemize} 
In each case, the proof by contradiction is similar to the proof in Case 2 above, where we pick a pair of $x_m, x_n$ and shows that $p^a_g$ can not achieve a lower value of the objective function than $p^*_g$. We thus do not repeat the argument here. QED

\newpage

\section{ \edit{Benefits of dual generator technique on 1D discrete example} }
\label{sec:1d_example}

\begin{figure}[h]
\centering
\includegraphics[width=\textwidth]{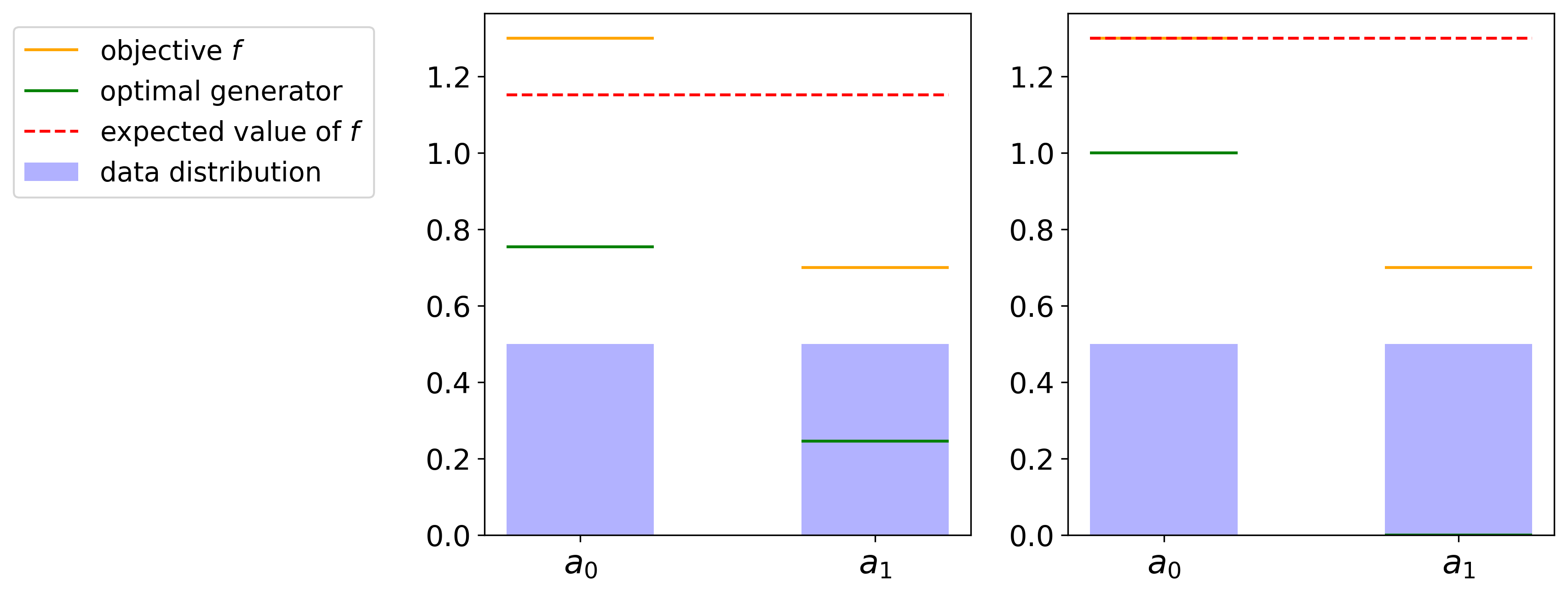}
\caption{\textbf{Left:} We use a single generator. \textbf{Right:} We use dual generator technique. The optimal generator (green bar) refers to the optimal primary generator, and not the auxiliary generator. Thanks to the dual generator technique, the optimal primary generator in the right figure assigns probability $1.0$ to the global maxima $a_0$ of the secondary objective function $f$. The optimal primary generator in the right figure is therefore better at maximizing the function $f$ than the optimal generator in the left figure. Note that in this example, the generator aims to maximize the function $f$ (instead of minimize) for more intuitive interpretation.}
\label{fig:1d_discrete_viz}
\end{figure}

In this section, we provide a simple one-dimensional numerical example with discrete action space to illustrate the benefit of the dual generator technique. In this example, the action space only consists two actions $a_0, a_1$. The probability of action $a_0$ under the data distribution is $0.5$. The probability of action $a_1$ under the data distribution is $0.5$. We would like to maximize a secondary objective function $f$. The function $f$ assigns value $1.3$ to $a_0$ and value $0.7$ to $a_1$.

We will next show in a self-contained Jupyter notebook that when using only a single generator, the expected value of the secondary objective function $f$ under the optimal generator is $1.15$. We also show that when using the dual generator technique, the expected value of the secondary objective function $f$ under the optimal primary generator is $1.3$. Since $1.3$ is higher than $1.15$, we can see that using the dual generator technique allows us to better maximize the objective function $f$.

Fig~\ref{fig:1d_discrete_viz} visually illustrates the benefits of the dual generator technique over using only a single generator in this example. We can also observe from Fig~\ref{fig:1d_discrete_viz} (Left) that the optimal generator is clearly different from the data distribution when only using a single generator. As such, in the GAN framework, the discriminator has an advantage in learning how to distinguish between samples from the real data distribution and samples from the generator. In contrast, when using the dual generator technique, the optimal primary generator assigns probability $1$ to action $a_0$ and the optimal auxiliary generator assigns probability $1$ to action $a_1$ (not shown in Fig~\ref{fig:1d_discrete_viz}). Their mixed distribution therefore assigns probability $0.5$ to either actions, matching the data distribution.

The Jupyter notebook in the next page illustrates the computations necessary to obtain the optimal solution for the generator.



In [1]:
```python
import numpy as np
from scipy.optimize import fsolve
```

In this toy example, the action space consists of two actions, denoted by $a_0$ and $a_1$. The probability of action $a_0$ under the data distribution is $0.5$. The probability of action $a_1$ under the data distribution is $0.5$. The data distribution is represented by the variable $p_{data}$ in the code cell below.

In [2]:
```python
p_data = np.array([
    0.5, 0.5
])
```

The secondary objective function that we would like to **maximize** is represented by $f$, where $f(a_0) = 1.3$ and $f(a_1) = 0.7$.

In [3]:
```python
f = np.array([
    1.3, 0.7
])
```

When using only a single generator and not employing the dual generator technique, we can write the closed form solution of the optimal generator as a function of the data distribution $p_{data}$, the secondary objective $f$, and a Lagrange variable $nu$ as follows. The optimal solution of the generator is represented by the variable $p\_g\_star$ in the code cell below:

In [4]:
```python
def get_p_star_single_generator(p_data, f, nu):
    
    # We use f instead of -f because
    # we are maximizing f in this example.
    exp = np.exp( f - nu )
    
    p_g_star = p_data * ( exp / ( 2 - exp ) )
    
    return p_g_star
```

Since the distribution induced by the optimal generator must be a valid distribution, we solve for the Lagrange variable $nu$ such that the probabilities that the optimal generator assigns to the two actions $a_0, a_1$ sum to 1.

In [5]:
```python
def func(nu):
    
    # p_g_star is a array consisting of two elements.
    # The first element is the density that the optimal
    # generator assigns to the action a_0.
    # The second element is the density that the optimal
    # generator assigns to the action a_1.
    p_g_star = get_p_star_single_generator(p_data, f, nu)
    
    sum_p_g_star = p_g_star.sum()
    
    return 1.0 - sum_p_g_star

# nu_star is the value of nu that makes the distribution
# induced by the optimal generator a valid distribution.
nu_star = fsolve(
    func, 1.
)
print('nu_star is', nu_star)
```

nu_star is [1.11566684]

In [6]:
```
# Confirm that the probabilities sum to 1
p_g_star = get_p_star_single_generator(p_data, f, nu_star)

print('The probabilities that the optimal generator assigns to the two actions are:')
print(p_g_star)

print('Sum of probabilities', p_g_star.sum().round(2))
```

```
The probabilities that the optimal generator assigns to the two actions are:
[0.75378698 0.24621302]
Sum of probabilities 1.0
```

In [7]:
```
E_f = (p_g_star * f).sum()
print(
    'Expected value of f under the optimal generator when using a single generator is:'
)
print(E_f.round(2))
```

```
Expected value of f under the optimal generator when using a single generator is:
1.15
```

We next derive the optimal solution of the generator when using the dual generator technique and show that using the dual generator technique allows us to better maximize the secondary objective function $f$. When using the dual generator technique, we maintain two generators, a primary generator denoted by $p_g$ and an auxiliary generator denoted by $p_{aux}$.

The in-support global maxima of $f$ is $a_0$, since $f(a_0) = 1.3 > f(a_1) = 0.7$. As such, the optimal solution of the primary generator, denoted by $p_g^*$, would like to assign as much probability to $a_0$ as possible. Since we have the constraint $p_g^*(a_0) \leq 2p_{data}(a_0) = 1.0$, we have that $p_g^*(a_0) = 1.0$. That is, the optimal primary generator assigns probability $1.0$ to action $a_0$. Since there are only two actions, the optimal primary generator assigns probability $0.0$ to the remaining action $a_1$.

The optimal primary generator is represented by the variable $dual\_p\_g\_star$ in the code cell below:

In [8]:
```
dual_p_g_star = np.array([
    1.0, 0.0
])

dual_E_f = (dual_p_g_star * f).sum()
print(
    'Expected value of f under the optimal generator when using dual generator is:')
print(dual_E_f.round(2))
```

```
Expected value of f under the optimal generator when using dual generator is:
1.3
```

Since we want to **maximize** the secondary objective function $f$, and $1.3$ is higher than $1.15$, we can see that using the dual generator technique allows us to better maximize the secondary objective function $f$.



\section{Description of the offline dataset generation procedure for the \texttt{noisy} and \texttt{biased} AntMaze datasets}

In the experiments section, we introduce the \texttt{bias} and \texttt{noisy} datasets for the AntMaze tasks. In this section, we provide more details on how the datasets were generated in the form of Python syntax in Code Listing~\ref{code-ant-dataset}. We plan to open-source both the datasets and the code to generate the datasets upon acceptance.

\begin{mypython}[caption={Illustration of the dataset generation procedure for the \texttt{bias} and \texttt{noisy} datasets. Given an \pyth{action} computed by the \pyth{behavior_policy}, we add noise and bias to the action. The magnitudes of the noise and bias depend on the x-values of the position of the Ant in the 2D maze.},label=code-ant-dataset]
NOISES = [0.1, 0.0, 0.2, 0.05, 0.3, 0.1, 0.4, 0.2]
BIASES = [0.1, -0.1, 0.2, 0.0, 0.2, -0.3, 0.2, 0.0]
POSITION = [-20.0, 0.0, 4.0, 8.0, 12.0, 16.0, 20.0, 24.0]

action = behavior_policy.get_action(obs)

x_position = get_x_position(obs)

pos = [idx for idx in range(len(POSITION)) if POSITION[idx] <= x_position]
pos = max(pos)

noise = NOISES[pos]
bias = BIASES[pos]

action = action + np.random.normal(size=action.shape) * noise - bias * np.ones_like(action)
action = np.clip(action, -1.0, 1.0)
\end{mypython}

\section{Additional experimental details}
\label{sec:additional_exp_details}

For all tasks, we average mean returns overs 20 evaluation trajectories. Similar to the pre-processing steps in previous works \cite{kostrikov2021offlineb}, we standardize MuJoCo locomotion task rewards by dividing by the difference of returns of the best and worst trajectories in each dataset. For the AntMaze datasets, we subtract 1 from rewards for all transitions. We use Adam optimizer \cite{adam} with a learning rate of $0.0003$. For the value functions, we use an MLP with $3$ hidden layers of size $256$. For both the GAN discriminator and auxiliary generator, we use an MLP with $1$ hidden layer of size $750$. The auxiliary generator takes a state as an input, and a noise vector and output actions deterministically as a function of the input state and noise vector. For the policy, which is also the primary generator, we use an MLP with $4$ hidden layers of size $256$. The policy takes a state as an input and outputs the parameters of a diagonal Gaussian, from which we sample an action. We update
the target network with soft updates with parameter $0.005$.

For the discriminator loss function, we use the mean-squared error loss, inspired by LSGAN \cite{lsgan}. For the auxiliary generator, we use the standard vanilla GAN loss. The loss functions and how they are used are further illustrated in Section~\ref{sec:detailed_algo_desc}. We also use instance noise \cite{gan_instance_noise} where we sample the instance noise from a Gaussian distribution for each action dimension independently. The Gaussian is zero-center and has an initial standard deviation of $0.3$ at the beginning of training. We anneal the magnitude of the noise over time and also clamp the instance noise to have a maximum magnitude of $0.3$. \edit{We also train the discriminator for more steps than the generators in each algorithm step (5 training steps for the discriminator for every step of the generator).}

In the policy objective (Eq.~\ref{eq:dasco_pi}), we also use a hyper-parameter $w$ to weight the contribution of the value function and the discriminator probability to the policy update. That is, we use Eq.~\ref{eq:dasco_pi_with_weight} to update the policy. We fix the value of $w$ throughout training. For the AntMaze tasks, we set $w=0.025$. For the Mujoco locomotation task, we set $w=1.0$. 
\begin{align}
    \policy^{k+1} \leftarrow& \argmax_{\policy} \E_{s, a_{\mathcal{D}} \sim \mathcal{D}, a \sim \policy^k(a|s)}\left[  \dfrac{1}{w} \dfrac{ D^k(s,a) }{ D^k( s,a_{\mathcal{D}}(s)) }Q^{k+1}(s, a) + \log D^k(s,a) \right],
    \label{eq:dasco_pi_with_weight}
\end{align}

In terms of total amount of compute and type of resources used, we use an internal cluster that allows for access up to 64 preemptive Nvidia RTX 2080 Ti GPUs. For each experiment of learning from an offline dataset, we use half a GPU and 3 CPU cores. Each experiment generally takes half a day to finish. We implemented our algorithms in Pytorch~\citep{paszke2019pytorch}. 

\section{\edit{Additional details on baselines}}
\label{sec:baseline_details}

\textbf{IQL} To obtain the result of IQL when learning from the \texttt{noisy} and \texttt{biased} AntMaze datasets presented in \autoref{tab:hetero_antmaze}, we tune the expectile hyper-parameter $\tau$ in IQL. \autoref{tab:iql_antmaze_newdataset_different_tau} illustrates the different values of $\tau$ we ran IQL for and the performance of each value.

\begin{table}[!htp]\centering
\caption{Performance on IQL when learning from the \texttt{noisy} and \texttt{biased} AntMaze datasets for different expectile value tau. tau=0.9 performs the best in the large maze, whereas tau=0.95 perform the best in the medium size maze. We therefore present the result using tau=0.9 in \autoref{tab:hetero_antmaze}. For lower tau values such as $0.8, 0.7$, the performance is worse. This is expected and also mentioned in Section 5.2 of the IQL paper.}\label{tab:iql_antmaze_newdataset_different_tau}
\small
\begin{tabular}{l || rrrrr }
Dataset & tau=1.0 & tau=0.95 & tau=0.9 & tau=0.8 & tau=0.7 \\ \hline
antmaze-large-bias & 0.0 (0.0) & 22.0 (0.8) & \textbf{41.0} (7.9) & 20.0 (3.6) & 4.0 (1.4)\\
antmaze-large-noisy & 0.0 (0.0) & 37.7 (5.2) & \textbf{39.0} (6.4) & 14.0 (5.9) & 3.3 (2.1)\\
antmaze-medium-bias & 0.0 (0.0) & \textbf{62.3} (9.8) & 48.0 (5.9) & 43.0 (7.1) & 21.0 (2.9)\\
antmaze-medium-noisy & 0.0 (0.0) & \textbf{53.7} (9.0) & 44.3 (1.7) & 41.7 (6.1) & 19.3 (2.5)\\
\hline
\end{tabular}
\end{table}

\textbf{CQL} We also tune the Lagrange threshold in CQL when learning from the \texttt{noisy} and \texttt{biased} AntMaze datasets. We refer to this hyper-parameter also as tau (but this is different from the hyper-parameter tau in IQL). The results for different values of the Lagrange threshold tau for CQL are illustrated in \autoref{tab:cql_antmaze_newdataset_different_tau}. 

\begin{table}[!htp]\centering
\caption{Performance of CQL when learning from the \texttt{noisy} and \texttt{biased} AntMaze datasets for different Lagrange threshold value tau. tau=2.0 performs the best overall and is the result we used in \autoref{tab:hetero_antmaze} for CQL.}\label{tab:cql_antmaze_newdataset_different_tau}
\small
\begin{tabular}{l || rrr }
Dataset & tau=1.0 & tau=2.0 & tau=0.5  \\ \hline
antmaze-large-bias & 50.0 (5.3)  & 61.7 (3.5) & 5.7 (3.1) \\
antmaze-large-noisy & 41.7 (4.6)  & 50.3 (2.3) & 5.0 (5.0) \\
antmaze-medium-bias & 72.7 (7.0)  & 66.7 (2.9)  & 31.7 (13.0)  \\
antmaze-medium-noisy & 55.0 (5.3) & 55.7 (4.7)  & 17 (3.6) \\
\hline
\end{tabular}
\end{table}

\textbf{EDAC} To obtain the results for EDAC presented in \autoref{tab:hetero_antmaze}, we performed a sweep over the number of value function used in the ensemble and the weight $\eta$ (used in step 5 in Algorithm 1 in EDAC paper to weight the gradient penalty term $ES$). For the number of value function, we ran the sweep over $20, 50, 100$. For the weight $\tau$, we ran the sweep over the values $1, 5, 10, 50, 100, 1000$. In other words, we ran $3 \times 6=18$ different combinations of hyper-parameter values for EDAC. We use the implementation of EDAC released by the original authors. Since we discovered that EDAC does not perform well in the medium size maze, we did not obtain the results of EDAC for the large maze. 

\textbf{BEAR} For BEAR, we used the laplacian kernel and performed a sweep over the hyper-parameter $mmd\_sigma$ using the values $1, 10, 20, 50$. We use the implementation available publicly at \url{https://github.com/rail-berkeley/d4rl_evaluations}. Similarly to EDAC, because BEAR does not perform well when learning from the medium size maze datasets, we did not obtain the results of BEAR for the large maze.

\section{Detailed algorithm description}
\label{sec:detailed_algo_desc}

Algorithm \autoref{algo:dasco} provides a summary of the training step given a batch of transitions from the offline dataset. In this section, we provide the description of how the different networks in our algorithms are trained using Python syntax. We include four Code Listings below, each illustrating the details of an update step in Algorithm \autoref{algo:dasco}.

\clearpage
\begin{mypython}[caption={Value networks training step given a batch of data, corresponding to step 4 in Algorithm \autoref{algo:dasco}},label=code-value-function-update]
rewards = batch['rewards']
terminals = batch['terminals']
obs = batch['observations']
actions = batch['actions']
next_obs = batch['next_observations']

# Computing target Q values
next_obs_target_actions = policy(next_obs)

target_Q1 = target_qf1(next_obs, next_obs_target_actions)
target_Q2 = target_qf2(next_obs, next_obs_target_actions)
target_Q = torch.min(target_Q1, target_Q2)
target_Q = rewards + (1 - terminals) * discount * target_Q

# Obtain loss function
current_Q1, current_Q2 = qf1(obs, actions), qf2(obs, actions)

qf1_loss = F.mse_loss(current_Q1, target_Q)
qf2_loss = F.mse_loss(current_Q2, target_Q)

# Update parameters of value functions
qf1_optimizer.zero_grad()
qf1_loss.backward()
qf1_optimizer.step()

qf2_optimizer.zero_grad()
qf2_loss.backward()
qf2_optimizer.step()

# Update Target Networks
soft_update_from_to(qf1, target_qf1, tau)
soft_update_from_to(qf2, target_qf2, tau)
\end{mypython}

\clearpage
\begin{mypython}[caption={Policy network training step given a batch of data, corresponding to step 5 in Algorithm \autoref{algo:dasco}},label=code-policy-function-update]
obs = batch['observations']
real_actions = batch['actions']

actor_actions = policy(obs)

# Compute value estimate
Q_pi_actions = qf1(obs, actor_actions)

# Compute log probability under discrimator
D_actor_actions_logit = discriminator(
    obs,
    actor_actions,
    return_logit=True
)

log_D_actor_actions = F.logsigmoid(D_actor_actions_logit)

# Compute probability ratio 
probs = discriminator(obs, actor_actions)
real_actions_probs = discriminator(obs, real_actions)

probs = torch.min(real_actions_probs, probs)

# min (D(s, a), D(s, a_dataset)) / D(s, a_dataset)
probs = probs / real_actions_probs

probs = probs.detach()

# Compute loss and update policy
policy_loss = - (
    probs * Q_pi_actions / w + log_D_actor_actions
).mean()

policy_optimizer.zero_grad()
policy_loss.backward()
policy_optimizer.step()

\end{mypython}

\clearpage
\begin{mypython}[caption={Auxiliary generator training step given a batch of data, corresponding to step 6 in Algorithm \autoref{algo:dasco}},label=code-aux-generator-function-update]
obs = batch['observations']

# Calculate loss
b_size = obs.size(0)
real_label = torch.full(
        (b_size,), 
        1) 

actions_fake = auxiliary_generator(obs)

logits = discriminator(
    obs,
    actions_fake,
    return_logit=True)

err = F.binary_cross_entropy_with_logits(
    logits, 
    real_label)

# Update auxiliary generator
auxiliary_generator_optimizer.zero_grad()
err.backward()
auxiliary_generator_optimizer.step()

\end{mypython}

\clearpage
\begin{mypython}[caption={Discriminator training step given a batch of data, corresponding to step 7 in Algorithm \autoref{algo:dasco}},label=code-discriminator-function-update]
obs = batch['observations']
actions = batch['actions']

b_size = obs.size(0)

# Calculate loss on real action
D_real_logits = discriminator(
    obs,
    actions + get_instance_noise(actions),
    return_logit=True
)

real_label = torch.full(
        (b_size,), 
        1) 

errD_real = F.mse_loss(
    F.sigmoid(D_real_logits), 
    real_label
) / 2. 

# Calculate loss on fake action 
def loss_fake_action(fake_action):
    fake_label = torch.full(
        (b_size,),
        0,
    )

    D_fake_logits = discriminator(
        obs,
        fake_action.detach() + get_instance_noise(fake_action),
        return_logit=True
    )

    errD_fake = F.mse_loss(
        F.sigmoid(D_fake_logits), 
        fake_label
    ) / 2. 
    
    return errD_fake
    
fake_action_aux = auxiliary_generator(obs)
fake_action_policy = policy(obs)

err_D_fake = loss_fake_action(fake_action_aux) \
    + loss_fake_action(fake_action_policy)

# Compute gradient and update the discriminator 
discriminator_optimizer.zero_grad()
(errD_real + err_D_fake).backward()
discriminator_optimizer.update()
\end{mypython}



\end{document}